\documentclass[journal]{IEEEtran}

\pdfoutput=1

\usepackage{cite}

\usepackage{ulem}

\ifCLASSINFOpdf
\else
\fi
\usepackage{algorithmic}
\usepackage{array}

\usepackage{url}
\usepackage{booktabs} 
\usepackage{times}
\usepackage{amsmath}
\usepackage{amssymb}
\usepackage{algorithm} 
\usepackage{algorithmic} 
\usepackage{enumerate} 
\usepackage{multirow} 

\usepackage{graphicx} 
\usepackage{subfigure} 
\usepackage{CJK} 
\usepackage{booktabs}
\usepackage{cite} 
\usepackage{color} 

\newcommand{\Smat}[0]{\ensuremath{{\bf S}} }

\newcommand{\Wmat}[0]{\ensuremath{{\bf W}} }
\newcommand{\Xmat}[0]{\ensuremath{{\bf X}} }
\newcommand{\Ymat}[0]{\ensuremath{{\bf Y}} }
\newcommand{\Zmat}[0]{\ensuremath{{\bf Z}} }

\newcommand{\av}[0]{\ensuremath{\boldsymbol{a}} }
\newcommand{\bv}[0]{\ensuremath{\boldsymbol{b}} }

\newcommand{\hv}[0]{\ensuremath{\boldsymbol{h}} }

\newcommand{\kv}[0]{\ensuremath{\boldsymbol{k}} }

\newcommand{\pv}[0]{\ensuremath{\boldsymbol{p}} }

\newcommand{\rv}[0]{\ensuremath{\boldsymbol{r}} }
\newcommand{\sv}[0]{\ensuremath{\boldsymbol{s}} }

\newcommand{\uv}[0]{\ensuremath{\boldsymbol{u}} }

\newcommand{\wv}[0]{\ensuremath{\boldsymbol{w}} }
\newcommand{\xv}[0]{\ensuremath{\boldsymbol{x}} }

\newcommand{\zv}[0]{\ensuremath{\boldsymbol{z}} }

\newcommand{\Pimat}[0]{\ensuremath{\boldsymbol{\Pi}} }

\newcommand{\Phimat}[0]{\ensuremath{\boldsymbol{\Phi}}}

\newcommand{\Omegamat}[0]{\ensuremath{\boldsymbol{\Omega}}}

\newcommand{\epsilonv}[0]{\ensuremath{\boldsymbol{\epsilon}} }

\newcommand{\etav}[0]{\ensuremath{\boldsymbol{\eta}} }

\newcommand{\lambdav}[0]{\ensuremath{\boldsymbol{\lambda}} }

\newcommand{\nuv}[0]{\ensuremath{\boldsymbol{\nu}} }

\newcommand{\piv}[0]{\ensuremath{\boldsymbol{\pi}} }

\newcommand{\sigmav}[0]{\ensuremath{\boldsymbol{\sigma}} }

\newcommand{\phiv}[0]{\ensuremath{\boldsymbol{\phi}} }

\newcommand{\omegav}[0]{\ensuremath{\boldsymbol{\omega}} }

\newcommand{\cdotv}[0]{\ensuremath{\boldsymbol{\cdot}}}
\newcommand{\mc}{\multicolumn}
\newcommand{\mr}{\multirow}

\newcommand{\given}{\,|\,}

\begin{document}
\title{Variational Temporal Deep Generative Model for Radar HRRP Target Recognition}
%
\author{Dandan~Guo,
 Bo~Chen,~\IEEEmembership{Senior Member,~IEEE,}
 Wenchao~Chen,
 Chaojie~Wang,
 Hongwei~Liu,~\IEEEmembership{Member,~IEEE,}
 and Mingyuan~Zhou

\thanks{
This research was partially supported by the Program for Oversea Talent by the Chinese Central Government, the 111 Project under Grant B18039, NSFC under Grant 61771361, the National Science Fund for Distinguished Young Scholars of China (61525105), and Shaanxi Innovation Team Project.
(Corresponding author: Bo Chen.)

Dandan Guo, Bo Chen, Wenchao Chen, Chaojie Wang and Hongwei Liu are with the National Laboratory of Radar Signal Processing, Xidian University, Xi'an 710071, China. (e-mail:gdd\_xidian@126.com; bchen@mail.xidian.edu.cn; wcchen\_xidian@163.com; xd\_silly@163.com; hwliu@xidian.edu.cn)
M. Zhou is with McCombs School of Business, The University of Texas at
Austin, Austin, TX 78712, USA. (e-mail: mingyuan.zhou@mccombs.utexas.edu)}}
%
\maketitle

\begin{abstract}
We develop a recurrent gamma belief network (rGBN) for radar automatic target recognition (RATR) based on high-resolution range profile (HRRP), which characterizes the temporal dependence across the range cells of HRRP. The proposed rGBN adopts a hierarchy of gamma distributions to build its temporal deep generative model. For scalable training and fast out-of-sample prediction, we propose the hybrid of a stochastic-gradient Markov chain Monte Carlo (MCMC) and a recurrent variational inference model to perform posterior inference.
To utilize the label information to extract more discriminative latent representations, we further propose supervised rGBN to jointly model the HRRP samples and their corresponding labels.
Experimental results on synthetic and measured HRRP data show that the proposed models are efficient in computation, have good classification accuracy and generalization ability, and provide highly interpretable multi-stochastic-layer latent structure.
\end{abstract}


\IEEEpeerreviewmaketitle

\section{Introduction}
\IEEEPARstart{R}adar automatic target recognition (RATR) is to identify the
unknown target from its radar echoes, which plays an important role in many applications,
such as surveillance, homeland security, and military
tasks~\cite{zyweck1996radar,bhanu1997guest,chiang2000model-based,sun2007adaptive,chen2009large,zhang2012multi-view,srinivas2014sar,ding2016convolutional,zhang2001a,du2005radar,YINGYANG2011,molchanov2014classification,Du2006A,Du2008Radar,feng2017radar,du2011bayesian,Pan2012Multi,Liao}.
Generally, the {researches} of RATR in high-resolution wideband radar can be divided into two categories, including RATR based on high-resolution range profile (HRRP) and that based on synthetic aperture radar (SAR) or inverse SAR (ISAR).
Compared with SAR and ISAR, HRRP owns the distinct advantage of being processed directly without first forming an image~\cite{du2011bayesian}.
Specifically, HRRP is composed of the amplitude of the coherent summations of the complex returns from target scatterers in each range cell, which represents the projection of the complex echoes from the target scattering center onto the radar line-of-sight (LOS).
Since HRRP contains abundant discriminative information, such as the target size and scatterer distribution, HRRP based RATR has received significant attention~\cite{zhang2001a,du2005radar,YINGYANG2011,molchanov2014classification,Du2006A,Du2008Radar,feng2017radar,du2011bayesian,Pan2012Multi,Liao}.

As the key step of HRRP based RATR, feature extraction has been widely studied, which can be motivated by different opinions according to the specific requirement.
For example, based on the integration of the bispectra of range profiles, Liao~et~al.~\cite{Liao} investigate bispectral features from real HRRP data.
Zhang~et~al.~\cite{zhang2001a} and Du~et~al.~\cite{du2005radar} further develop HRRP target recognition methods using high-order spectra features for their shift-invariance properties.
Molchanov~et~al.~\cite{molchanov2014classification} study the possibility of classifying aerial targets using the micro-Doppler signatures, where the novel features are computed in the form of cepstral coefficients and bicoherence estimates. Despite their usefulness for target recognition, those engineered features are hand-crafted and heavily rely on personal experiences, limiting their use in practice
~\cite{feng2017radar}.

To learn data-driven features, Du~et~al.~\cite{Du2007Radar} introduce principal component analysis (PCA) to extract the complex HRRPs' feature subspace, by minimizing the reconstruction error for the HRRP data. Du~et~al.~\cite{Du2012Radar} further propose a factor analysis (FA) model based on multitask learning to describe the Fast Fourier Transform (FFT)-magnitude feature of complex HRRP.
Some researchers~\cite{Feng2012Radar} adopt the K-Singularly Valuable Decomposition (K-SVD) dictionary learning method to extract the desired sparse over-complete features of HRRP data. Those methods have proven their effectiveness in practice, but they are all shallow models and only good at extracting linear features~\cite{du2019factorized}.
Inspired by the ability of deep learning methods in extracting multilayer non-linear features~\cite{Lecun2015Deep}, Feng~et~al.~\cite{feng2017radar} adopt stacked corrective auto-encoders (SCAE) to extract robust hierarchical features for HRRP, employing the average profile of each HRRP frame as the correction term.
Pan~et~al.~\cite{pan2017radar} propose a discriminant deep belief network (discriminant DBN) to recognize non-cooperative targets with imbalanced HRRP training data.
Nevertheless, all these models only depict the global structure of the target in a single HRRP sample, ignoring the sequential relationships across range cells within a single HRRP sample.

Several approaches have been proposed to exploit the temporal dependence in HRRP.
Du~et~al.~\cite{du2011bayesian} propose a Bayesian dynamic model for the HRRP sequence, where the spatial structure across range cells in HRRP is depicted by a hidden Markov model (HMM) and the temporal dependence between HRRP samples is depicted by the time evolution of the transition probabilities.
Pan~et~al.~\cite{Pan2011Multi,Pan2012Multi} characterize the spectrogram feature from a single HRRP sample via an HMM, which is a two-dimensional time-frequency representation providing the variation of the target in both the frequency and time domains.
Besides, Wang~et~al.~\cite{WangLDS} characterize the frequency spectrum amplitude of HRRP based on linear dynamical systems (LDS) to relax the requirement of a large training set.
Despite the tremendous success of HMM and LDS in many areas, efficiently capturing complex dependencies between range cells in HRRP remains a challenging problem due to their limited expressiveness. With the development of temporal one-dimensional convolution neural network (CNN) in raw audio generation tasks \cite{denoord2016wavenet}, Wan~et~al.~\cite{wan2019convolutional} use one-dimensional CNN to handle HRRP in the time domain and construct a two-dimensional CNN model for the corresponding spectrogram representation. Besides, recurrent neural networks (RNNs)~\cite{elman1990finding,graves2013speech,XuBin} have been developed to capture complex temporal behavior in high-dimensional sequences. While in principle RNN is a powerful model, it does not consider the kind of variability observed in highly structured data~\cite{PascanuICML2013,chung2015a} and ignores the weight uncertainty when updating its parameters with stochastic optimization such as stochastic gradient descent (SGD)~\cite{gan2017scalable}.
Moreover, learning both hierarchical and temporal representations has been a long-standing challenge for RNNs, in spite of the fact that hierarchical structures naturally exist in many sequential data and the learned latent hierarchical structures can provide useful information to other downstream tasks such as sequence generation and classification~\cite{chung2017hierarchical,koutnik2014a}.
In addition, the neural network representations are generally inaccessible and uninterpretable to humans~\cite{Zaheer2017a}.
More recently, several deep probabilistic dynamical models have been proposed to capture the relationships between latent variables across
multiple stochastic layers on video and music sequences, text streams, and motion capture data~\cite{gan2015deep,gong2017deep,guo2018deep}.

Deep Poisson gamma dynamical system ({DPGDS}) is a deep Bayesian top-down directed generative model (decoder) ~\cite{guo2018deep}, which takes advantage of the hierarchical structure to efficiently incorporate both between-layer and temporal dependencies. Different from classical Kalman filters, such as LDS, where the uses of linear transition and emission distribution limit the capacity to model complex phenomena \cite{krishnan2015deep}, DPGDS directly chains the observed data in a state space model (deep gamma Markov chain) that evolves with gamma noise. To take advantage of the temporal dependence within each HRRP sequence, we can construct a sequential HRRP RATR model with DPGDS, where each HRRP can be divided into multiple overlapping sequential HRRP segments as input. Despite being able to infer the multilayer contextual representation of {observed HRRP sequences} with scalable inference, the inference of {DPGDS} relies on a potentially large number of Markov chain Monte Carlo (MCMC) iterations to extract the latent representation of a new sample at the testing stage, which may be unattractive when real-time processing is desired. {Thus, the key challenge for DPGDS in RATR (unconventional deep Markov chain) is to infer the gamma distributed latent states with higher efficiency.} In this paper, we first generalize  {DPGDS} to recurrent gamma belief network (rGBN) for real-time processing at the test stage, by equipping it with a novel inference network (encoder).
Specifically, to provide scalable training and fast out-of-sample prediction, the potential solution is to build an inference model with a variational auto-encoder (VAE)~\cite{kingma2014autoencoding}. But existing VAE variants are mostly restricted to non-sequential observed data and Gaussian distributed latent variables.
To address these constraints, motivated by related constructions in Zhang et al.~\cite{Zhang2018WHAI}, we construct a novel recurrent variational inference network (encoder) to map the observed HRRP samples directly to their latent temporal representations.
Then we provide the hybrid of stochastic-gradient-MCMC (SG-MCMC) and recurrent variational inference to infer both the posterior distribution (rather than a point estimate) of the global parameters of generative model and latent temporal representations.
To the best of our knowledge, the proposed rGBN, characterized by a top-down generative structure with temporal feedforward structure on each layer and a novel inference model, is the first deep probabilistic dynamical model for the HRRP RATR task. Although the features unsupervisedly learned by {rGBN} can be fed into a downstream classifier to make predictions, it is often beneficial to incorporate the target label information into the model ~\cite{li2015maxmargin}. To explore this potential, we further develop an end-to-end supervised rGBN (s-rGBN), whose extracted features are good for both classification and data representation.

The remainder of the paper is organized as follows. In Section~II, we present a brief description of HRRP, HMM, LDS, and RNN. We introduce our generative and inference networks in Section~III.
By jointly modeling HRRP samples and their labels, we further propose the supervised model, $i.e.$, {s-rGBN}, in Section~IV.
The detailed experimental results based on synthetic data and measured
HRRP data are reported in Section~V.
Section VI~concludes the paper.

\section{Preliminaries}
\subsection{Review of HRRP}
\begin{figure}[!htbt]
\vspace{-0.1cm}
\setlength{\abovecaptionskip}{0.cm}
\setlength{\belowcaptionskip}{-2.cm}
\begin{center}
\centering
\includegraphics[height=3.5cm,width=4.2cm]{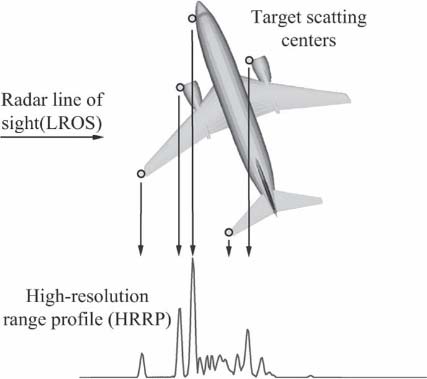}
\caption{Radar returns from the scatterers on the target are projected onto the
line-of-sight (LOS), resulting in an HRRP. This figure is quoted from Zwart et al. \cite{2003Zwart}.}
\label{fig:HRRP sample}
\end{center}
\end{figure}

For a high resolution radar (HRR), the wavelength of the radar signal is far smaller than the target size.
Intuitively, HRRP is a representation of the time domain response of the target to an HRR pulse as a one-dimensional signature, which is the expression of the distribution of radar scattering centers along with the radar LOS~\cite{du2011bayesian}\cite{Du2012Radar}\cite{du2019factorized}, as shown in Fig.~\ref{fig:HRRP sample}. Suppose the transmitted signal of an HRR is $s(t)e^{j2 \pi f_c t}$, where $s(t)$ denotes the complex envelop and $f_c$ is the radar carrier frequency. Following the literatures~\cite{Du2006A,Du2007Radar}, the $n$th complex echo from the $d$th range cell ($n=1,2,...,N,d=1,2,...,D$) in the baseband is defined as follows
\begin{gather}\label{hrrp}
x_d(t,n) = \sum_i^{P_d} \sigma_{di} s\left(t-\textstyle \frac{2R_{di}(n)}{c}\right)e^{-j [\frac{4\pi}{\lambda}R_{di}(n)-\theta_{di0}] }
\end{gather}
where $P_d$ represents the number of target scatterers in the $d$th range cell, $\lambda$ is the wavelength of HRR, $\sigma_{di}$ and $\theta_{di0}$ represent the intensity and initial phase of the $i$th scatterer in the $d$th range cell, respectively.
$R_{di}(n)$ can be regarded as the radial distance between the $i$th scatterer in the $d$th range cell of the $n$th returned echo and the radar.

We set $R(n)$ as the radial distance between the target reference center in the $n$th returned echo and the radar, and set $L_{x}$ as the radial length of the target, usually $R(n)\gg L_{x}$.
Due to the target rotation, there are radial displacements for the scatterers.
Then $R_{d i}(n)\approx R(n)+\Delta r_{d i}(n)$, where $\Delta r_{d i}(n)$ represents the radial displacement of the $i$th scatterer in the $d$th range cell in the $n$th returned echo. When $s(\cdot)$ is a rectangular pulse signal with unit intensity, it is usually omitted. Equation \eqref{hrrp} can be approximated as $x_{d}(t, n) \approx x_{d}(n)= \sum_{i}^{P_{d}} \sigma_{d i} e^{-j \frac{4 \pi}{\lambda} R(n)} e^{j \phi_{d i}(n)}$, where $\phi_{d i}(n)=\theta_{d i 0}-\frac{4 \pi}{\lambda} \Delta r_{d i}(n)$ denotes the remained echo phase of the $i$th scatterer in the $d$th range cell of the $n$th returned echo, and $e^{-j \frac{4 \pi}{\lambda} R(n)}$ represents the initial phase of the $n$th returned echo related to the target distance and radar wavelength. Since $e^{-j \frac{4 \pi}{\lambda} R(n)}$ does not contain the target discriminative information, we can eliminate it and define the $n$th real HRRP sample $\xv(n)$ as
\begin{equation}\label{real_HRRP}
\small{
\begin{aligned}
&\xv(n) \triangleq \left[|x_{1}(n)|, |x_{2}(n)|, \ldots, |x_{D}(n)|\right] \\
&=\small \left[\left|\sum_{i}^{P_{1}} \sigma_{1 i} e^{j \phi_{1 i}(n)}\right|,\left|\sum_{i}^{P_{2}} \sigma_{2 i} e^{j \phi_{2 i}(n)}\right|, \ldots\left|\sum_{i}^{P_{D}} \sigma_{D i} e^{j \phi_{D i}(n)}\right|\right],
\end{aligned}}
\end{equation}
where $|\cdot|$ means taking absolute value, and $D$ is the dimensionality of $\xv(n)$.

{\subsection{Related dynamical models}}
Denoting a sample as $T$ sequentially observed $V$-dimensional vectors, we present a short review of the existing dynamical models used to model the temporal dependence across the range cells in HRRP \cite{du2011bayesian,Pan2012Multi,Pan2011Multi,WangLDS,XuBin}.

\textbf{Hidden Markov Model:} The joint likelihood of the observation and underlying state sequence can be expressed as
\begin{gather}
P(\xv_{n},\sv_n\given\omegav_0,\Pimat,\phiv)=P(\sv_{1,n}\given\omega_0)\prod_{t=1}^{T-1}P(\sv_{t,n}\given\sv_{t-1,n},\Pimat) \notag \\
\times \prod_{t=1}^{T}P(\xv_{t,n}\given\sv_{t,n},\phiv),
\end{gather}
where $\sv_{n}=\{\sv_{1,n},.., \sv_{t,n},..,\sv_{T,n}\}$ denotes the latent states of the $n$th sample, $\omegav_0$ is the initial state probability, $\Pimat$ is the state transition distribution and $\phiv$ are a set of parameters governing the emission probability.

\textbf{Linear Dynamical System:}  It consists of the following observation and state equations
\begin{gather}
\xv_{t,n}\sim N(\Phimat\sv_{t,n},\Sigma),~\sv_{t,n}\sim N(\Pimat\sv_{t-1,n}, \Delta ),
\end{gather}
where $\Pimat$ is the transition matrix, $\Sigma$ and $\Delta$ are covariance matrices, and $\sv_{t,n}$ denotes the latent state that is projected to the observed space via the factor loading matrix $\Phimat$.

\textbf{Recurrent Neural Network:} At timestep $t$, RNN reads the symbol $\xv_{t,n}$ and updates its hidden state $\hv_{t,n}^{(l)}$ at layer $l$
\begin{equation}\label{RNN}
\small
\hv_{t,n}^{(l)}=\left\{\begin{array}{ll}{f \left(\Wmat_{hh}^{(l)}\hv_{t-1,n}^{(l)} + \Wmat_{xh}^{(l)}\xv_{t,n}\right)} ,& {\text { if } l=1 } ,\\
f\left(\Wmat_{hh}^{(l)}\hv_{t-1,n}^{(l)}+\Wmat_{xh}^{(l)}\hv_{t,n}^{(l-1)} \right) , & {\text { if } L\geq l>1}, \\
\end{array}\right.
\end{equation}
where $f$ is a deterministic non-linear function, $\Wmat_{hh}^{(l)}$ is the transition matrix, $\Wmat_{xh}^{(l)}$ is the weight matrix at layer~$l$, and the bias vectors are omitted for conciseness.

Although HMM and LDS have been widely studied, their representation power may be limited when modeling the HRRP sequential samples. Compared with LDS and HMM, RNN typically owns extra expressive power due to the existence of deep hidden states and flexible non-linear transition function. However, the internal structure of RNN is in general entirely deterministic, with the only source of variability provided by its conditional output probability model, which may be inappropriate to model the kind of variability observed in the HRRP data~\cite{chung2015a}.

\vspace{3mm}
\section{Variational Temporal Deep Generative Model}\label{my_model}

\subsection{Input representation}
\begin{figure}[!htbt]
\setlength{\abovecaptionskip}{-0.1cm}
\setlength{\belowcaptionskip}{-0.2cm}
\begin{center}
\centering
\includegraphics[height=3.2cm,width=8.1cm]{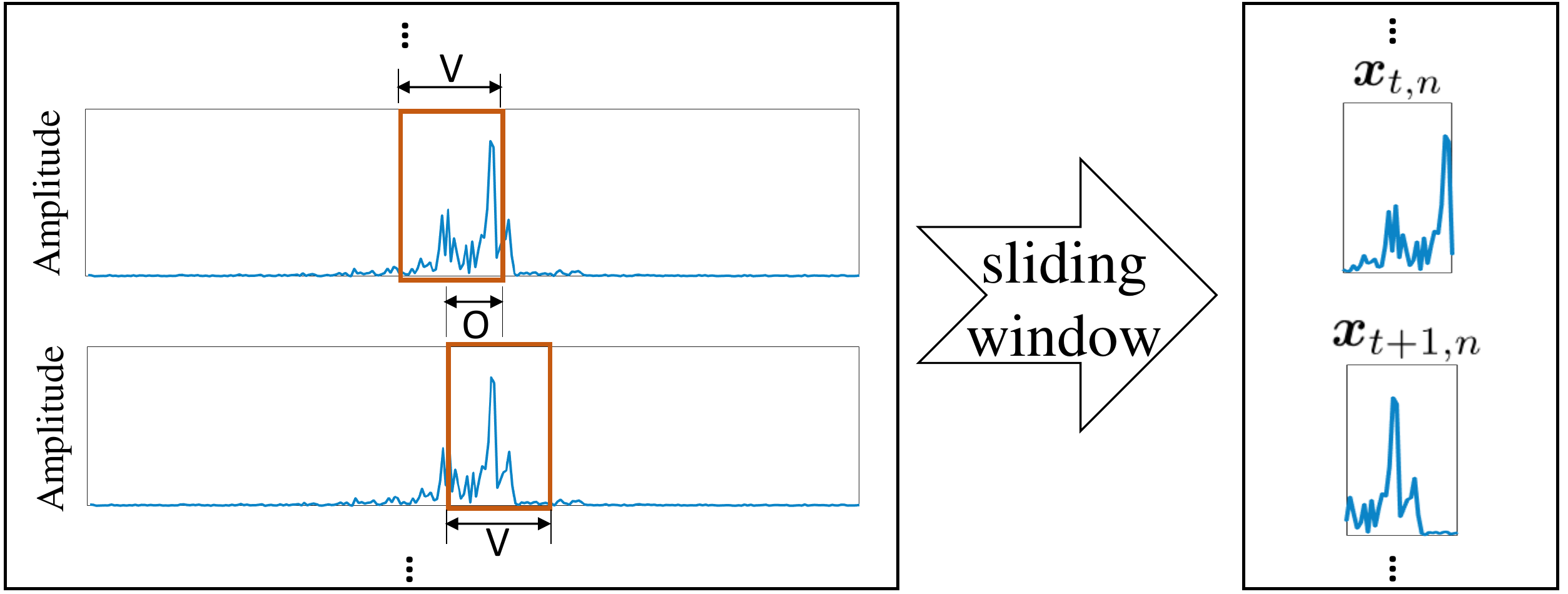}
\caption{ Visualization of the $n$th real HRRP sample $\xv(n) \in \mathbb{R}_{+}^{D} $ (left) and its corresponding time sequential features $\xv_n \in \mathbb{R}_{+}^{V \times T}$ (right), where $V$ represents the length of window, $O$ denotes the length of overlap of the window, $\xv_{t,n}$ denotes the input of $n$th HRRP sequence at timestep $t$.}
\label{fig:sequential feature of HRRP}
\end{center}
\end{figure}

According to~Xu et al. \cite{XuBin}, to discover the temporal dependence between the range cells within the single HRRP, we divide the $n$th $D$-dimensional real HRRP $\xv(n)$ in \eqref{real_HRRP} into $\xv_n\in \mathbb{R}_{+}^{V \times T}$, which consists of $T$ sequentially observed $V$-dimensional vector. Shown in Fig. \ref{fig:sequential feature of HRRP}, we denote the HRRP sequence as $\xv_n=[\xv_{1,n},..,\xv_{t,n},..\xv_{T,n}]$, where $\xv_{t,n}\in \mathbb{R}_{+}^{V \times 1}$ is the time sequential feature at timestep $t$ of  sample $n$ and can be defined as
\begin{align}\label{temporal_HRRP}
{\xv_{t,n} = \xv(n)\left( a_t+1 : a_t+V \right),}
\end{align}
where {$V$ is the size of window function, intercepting $V$ range cells from the real HRRP sample}, $a_t=(V \!-\! O)\ast(t \!-\! 1)$ is the index at timestep $t$, and {$O$ denotes the overlap length across the windows, determining the degree of correlation between adjacent timesteps. Thus $T = \left\lfloor {(D-O)/(V-O)} \right\rfloor$}. {We denote $\Xmat=\{\xv_n\}_{n=1}^{N}$ as the HRRP dataset, which consists of $N$ independent and identically distributed (IID) HRRP sequences, and $\Ymat=\{y_n\}_{n=1}^{N} $ as corresponding labels, where $y_n$ is the label of $\xv_n$. Note sequential inputs $\xv_{1:T,n}$ from one HRRP sample $\xv_n$ share the same label. The dataset $\{\Xmat,\Ymat\}$ can be fed into a dynamic model as $N$ IID samples, each of which contains $T$ sequentially observed $V$-dimensional vectors (with identical label). For simplicity, we only exhibit} the modeling process for the $n$th HRRP sequence.

\vspace{-0.1cm}
\vspace{-0.2cm}

\subsection{Generative Model}
\begin{figure*}[!htbt]
\begin{center}
\setlength{\abovecaptionskip}{-0.2cm}
\setlength{\belowcaptionskip}{-0.4cm}
\centering
\includegraphics[height=4.8cm,width=18cm]{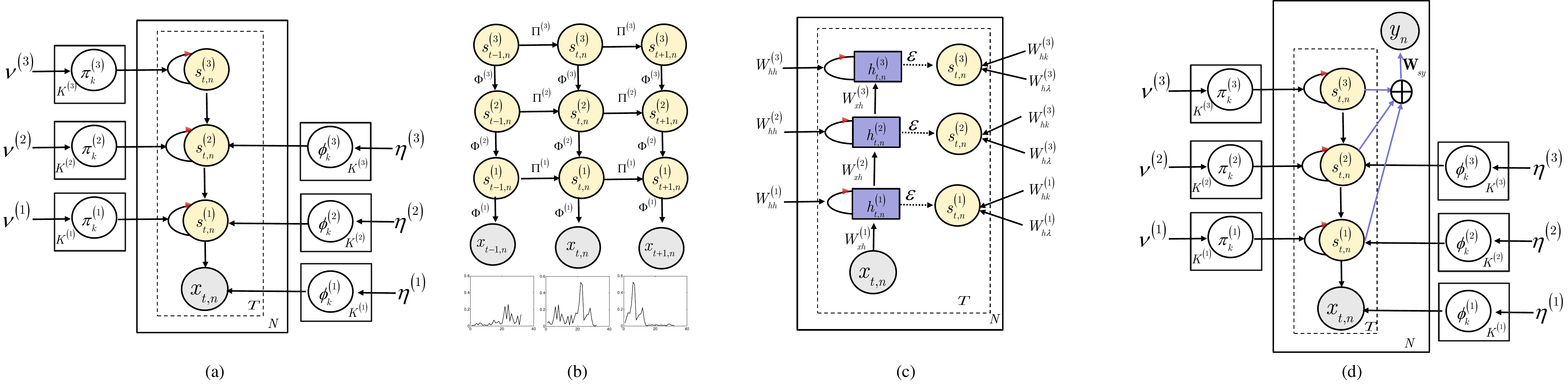}
\caption{(a) The generative model of recurrent gamma belief network (rGBN); (b) The unfolded generative model of rGBN for the $n$th HRRP sample; (c) The recurrent variational inference model of rGBN (ignoring all bias terms); (d) The generative model of supervised recurrent gamma belief network (s-rGBN). }
\label{fig:deep_dynamical_model}
\end{center}
\end{figure*}

To characterize the sequential feature within a single HRRP sample, we generalize the deep Poisson gamma dynamical system (DPGDS) \cite{guo2018deep} to rGBN, whose generative model is sketched in Fig.~\ref{fig:deep_dynamical_model}~(a). Specifically, we consider the deep architecture with $L$ layers, and denote $ \sv_{t,n}^{(l)}\in\mathbb{R}_+^{K_l}$ as the latent state of $\xv_{t,n}$ in \eqref{temporal_HRRP} at layer $l$, time step $t$, where $K_l$ is the number of states at layer $l$.
Different from HMM and LDS whose single-layer latent state at time step $t$ only relies on the state at previous time step $t-1$,
the latent state $ \sv_{t,n}^{(l)}$ of {rGBN}, from the top layer to bottom, is formulated as
\begin{align}\label{DPGDS_hidden}
\small
\sv_{t,n}^{(l)} \sim \mbox{Gam}\left(b^{(l)}\left( \Pimat^{(l)} \sv_{t-1,n}^{(l)} + \Phimat^{(l+1)} \sv_{t,n}^{(l+1)}\right) , 1/b^{(l)} \right),
\end{align}
where $x\sim \mbox{Gam}(a,1/b)$ denotes the gamma distribution with shape $a$, scale $1/b$, and mean $a/b$; $\Pimat^{(l)}\in \mathbb{R}_{+}^{K_{l} \times K_{l}}$ the transition matrix of layer $l$, $\Phimat^{(l)}\in \mathbb{R}_{+}^{K_{l-1} \times K_{l}}$ the weight matrix connecting layers $l-1$ and $l$, {$K_{0}=V$}, and $1/b^{(l)}$ the gamma scale parameter at layer $l$. When $t=1$, $\sv_{1,n}^{(l)} \sim \mbox{Gam}\left(b^{(l)}\Phimat^{(l+1)} \sv_{1,n}^{(l+1)} , 1/b^{(l)} \right)$ for $1\leq l< L$ and
$\sv_{1,n}^{(L)} \sim \mbox{Gam}\left(b^{(L)}\textrm{1}_{K_L} , 1/b^{(L)} \right)$. When $t>1$, $\sv_{t,n}^{(L)} \sim \mbox{Gam}\left(b^{(L)}\Pimat^{(L)} \sv_{t-1,n}^{(L)} , 1/b^{(L)} \right)$.

The gamma shape parameter of $\sv_{t,n}^{(l)}$ can be divided into two parts. One is the product of the transition matrix $\Pimat^{(l)}$ and latent state $\sv_{t-1,n}^{(l)}$, capturing the temporal dependence at the current layer, while the other is the product of the connection weight matrix $\Phimat^{(l+1)}$ and latent state $\sv_{t,n}^{(l+1)}$, capturing the hierarchical dependence at the current time. Moving beyond RNN using deterministic non-linear transition functions, we construct a dynamic probabilistic model using gamma distributed non-negative hidden units. Therefore, the proposed model is characterized by its expressive structure, which not only captures the correlations between the latent states across all layers and time steps, but also models the variability of the latent states, improving its ability to model sequential HRRP.

As a generative model with the deep temporal structure, the observed HRRP sequence at time step $t$ can be drawn from $p(\xv_{t,n}\given\Phimat^{(1)},\sv_{t,n}^{(1)})$.
To consider the non-negative nature of the HRRP data and facilitate inference, we introduce the Poisson randomized gamma (PRG) distribution defined as $\mbox{PRG}\left(\xv_{t,n}\given\Phimat^{(1)} \sv_{t,n}^{(1)},c\right)$ in Zhou et al.~\cite{Zhou2016JMLR}, which has a point mass at $\xv_{t,n}=0$ and is continuous for $\xv_{t,n}>0$. Since the PRG distribution is generated as a Poisson mixed gamma distribution, the data likelihood can be expressed as
\begin{gather}\label{DPGDS_sample_PRG}
\xv_{t,n} \sim \mbox{Gam}\left(\rv_{t,n},1/c\right), \rv_{t,n}\sim \mbox{Pois}\left(\Phimat^{(1)} \sv_{t,n}^{(1)} \right),
\end{gather}
where $c>0$, $ x\sim \mbox{Pois}(\lambda)$ represents the Poisson distribution with mean $\lambda$ and variance $\lambda$, and $\Phimat^{(1)}\in \mathbb{R}_{+}^{V \times K_{1}}$ is the weight matrix of layer $1$.

For scale identifiability and ease of inference and interpretation, we place the Dirichlet distribution prior on each column of $\Phimat^{(l)}$ and $\Pimat^{(l)}$ , $i.e.$, $\phiv_k^{(l)}$ and $\piv_k^{(l)}$, by letting
\begin{gather}\label{Prior}
\phiv_k^{(l)}\sim \textrm{Dir}(\etav^{(l)},...,\etav^{(l)}) ,\\
\piv_k^{(l)}\sim \textrm{Dir}(\nuv^{(l)},...,\nuv^{(l)}) ,
\end{gather}
for $l \in {1,...,L}$, which makes the elements of each column be non-negative and sum to one. Note $\piv_k^{(l)} =(\piv_{1,k}^{(l)},..,\piv_{k_1,k}^{(l)},..,\piv_{K_l,k}^{(l)}) $ and $\piv_{k_1,k}^{(l)}$ describes how much the weight of state $k$ of the previous time at layer $l$ is transited to influence state $k_1$ of the current time at the same layer. Under the hierarchical dynamical model defined by \eqref{DPGDS_hidden} and \eqref{DPGDS_sample_PRG}, the joint likelihood of the observation HRRP and the temporal latent states can be constructed as
\begin{align}\label{probability}
&P\left(\xv_{t,n},\{\sv_{t,n}^{(l)}\}_{l=1}^{L} \given \{\Phimat^{(l)},\Pimat^{(l)},b^{(l)}\}_{l=1}^{L}, c \right) \notag\\
&=\left[\prod\limits_{l = 1}^{L} {p\left(\sv_{t,n}^{(l)}\given\Phimat^{(l + 1)}\sv_{t,n}^{(l+ 1)},\Pimat^{(l)}\sv_{t-1,n}^{(l)}, b^{(l)} \right)} \right]\\
& \times p\left(\xv_{t,n}\given\Phimat^{(1)}\sv_{t,n}^{(1)},c \right) . \notag
\end{align}

The parameters of the generative model comprise the transition and weight matrices, which we write as {$\{\Pimat^{(l)},\Phimat^{(l)}\}_{l=1}^{L}$}.
Fig. \ref{fig:deep_dynamical_model} (a) shows the graphical representation of the proposed generative model and Fig. \ref{fig:deep_dynamical_model}(b) is the unfolded representation of the model structure.

\textbf{Structure Analysis:}
As discussed above, if $x\sim \mbox{Gam}(a,1/b)$, the mean of $x$ is $a/b$; while if $x \sim \mbox{Pois}(\lambda)$, the mean of $x$ is $\lambda$.
Therefore, the expected value of $\xv_{t,n}$ in \eqref{DPGDS_sample_PRG} and $\sv_{t,n}^{(l)}$ in \eqref{DPGDS_hidden} can be expressed as
\begin{gather}\label{expected_observed}
\small \mathbb{E}\left[\xv_{t,n}\given \rv_{t,n},c\right] = \mathbb{E}\left[\rv_{t,n}\given\Phimat^{(1)} \sv_{t,n}^{(1)}\right]/c= \Phimat^{(1)}\sv_{t,n}^{(1)}/c ,
\end{gather}
\begin{align}\label{expected_latent1}
 \small \mathbb{E}\left[\sv_{t,n}^{(l)} \given \sv_{t,n}^{(l+1)},\sv_{t-1,n}^{(l)} ,\Phimat^{(l+1)} , \Pimat^{(l)} \right]
 = \Phimat^{(l+1)}\sv_{t,n}^{(l+1)}+ \Pimat^{(l)}\sv_{t-1,n}^{(l)} .
\end{align}
Based on \eqref{expected_observed} and \eqref{expected_latent1}, for a three-hidden layer
rGBN shown in Fig. \ref{fig:deep_dynamical_model}(a), we have
\begin{align}\label{deep_temporal}
 &\mathbb{E} [\xv_{t,n}\,\given\, \sv_{t-1,n}^{(1)}, \sv_{t-2,n}^{(2)}, \sv_{t-3,n}^{(3)}]/c \notag\\
 &= \Phimat^{(1)} \Pimat^{(1)} \sv_{t-1,n}^{(1)}
 + \Phimat^{(1)} \Phimat^{(2)} [\Pimat^{(2)}]^{2} \sv_{t-2,n}^{(2)}\\
 &~~~~+ \Phimat^{(1)} \Phimat^{(2)}\left( \Pimat^{(2)}\Phimat^{(3)}+\Phimat^{(3)}\Pimat^{(3)}\right) [\Pimat^{(3)}]^{2} \sv_{t-3,n}^{(3)} , \notag
\end{align}
where the expected value of $\xv_{t,n}$ depends on the latent states at the previous time step in each layer, indicating our proposed model captures and transmits long-range temporal information through its higher hidden layers. In addition, the proposed model can be viewed as the generalization of LDS and HMM with
deep gamma distributed latent representations, and also can be considered as a probabilistic
construction of the traditionally deterministic RNN by adding uncertainty into the latent space via a deep generative model.

Ignoring the temporal structure of the equation, we notice that the information of the whole HRRP data set can be compressed into the inferred network $\{\Phimat^{(1)},\Phimat^{(2)}, \Phimat^{(3)}\}$, which depicts the global structure of the target in a single HRRP.
To be more specific, we can visualize the $\phiv_k^{(l)}$ at layer $l$ as $\left[\prod_{p=1}^{l-1}\Phimat^{(p)}\right]\phiv_k^{(l)}$,
which are often quite specific at the bottom layer and become increasingly more general when moving upwards. We will examine the weights of different layers to understand the general and specific aspects of the HRRP data, and illustrate how the weights of different layers are related to each other.

\textbf{Quantizing the HRRP data:}
While the non-negative real HRRP dataset can be modeled by PRG distribution, the latent count $\rv_{t,n}$ in \eqref{DPGDS_sample_PRG} need to be sampled in each iteration of the training stage. Specifically, the conditional posterior of $\rv_{t,n}$ given $\xv_{t,n}$, $\Phimat^{(1)} \sv_{t,n}^{(1)}$, and $c$ can be expressed as
\begin{align}
\resizebox{.88\hsize}{!} {$ p\left(\rv_{t,n}\given \xv_{t,n},\Phimat^{(1)} \sv_{t,n}^{(1)},c \right) \!= \! \frac{\mbox{Gam}\left(\rv_{t,n},1/c\right) \mbox{Pois}\left(\Phimat^{(1)} \sv_{t,n}^{(1)} \right)}{\sum \limits_{\rv_{t,n} \!= \! 0}^{\infty} \mbox{Gam}\left(\rv_{t,n},1/c\right) \mbox{Pois}\left(\Phimat^{(1)} \sv_{t,n}^{(1)} \right)} $} .
\end{align}
More details about the PRG distribution can be found in Zhou et al.~\cite{Zhou2016JMLR} and are omitted here for brevity.
Despite the desired ability to depict non-negative continuous data, the PRG distribution may be time-consuming for the iterative procedure to infer latent counts in real-world applications. Considering the limited computation power in the training stage, HRRP data can be modeled with Poisson distribution, by directly discretizing
the HRRP data to counts ($i.e.$, non-negative integers) before training, where the sampling step of the counts in each iteration can be avoided.
Therefore, instead of modeling the observed data with \eqref{DPGDS_sample_PRG}, we assume that
\begin{gather}\label{Poisson_observed}
\lfloor \mu \xv_{t,n}\rfloor \sim \mbox{Pois}\left(\Phimat^{(1)} \sv_{t,n}^{(1)} \right),
\end{gather}
where the scaling factor $\mu$ controls the fineness of this discretization.
There is a trade-off between the benefit of the PRG distribution and restriction of computation resources.
Note that modeling the observed HRRP  with the Poisson distribution can still obtain similar temporal dependencies in \eqref{deep_temporal} and hierarchical structure $\prod_{p=1}^{l}\Phimat^{(p)}$, omitted here for brevity.
\vspace{-0.2cm}
\subsection{Hybrid Inference Model} \label{my_inference}
In this section, we first infer the generative model global parameters $\{\Pimat^{(l)} \}_{l=1}^{L}$ and $\{ \Phimat^{(l)} \}_{l=1}^{L}$ with MCMC, then introduce a recurrent inference model to infer the latent states $\{ \sv_{t,n}^{(l)} \}_{t=1,l=1,n=1}^{T,L,N}$.
Finally, we provide a hybrid SG-MCMC and recurrent variational inference, which is scalable at the training stage and fast at the testing stage.

\textbf{MCMC inference for the generative network:}
Given the HRRP sequences, the inference task here is to find the weight matrices $\{ \Phimat^{(l)} \}_{l=1}^{L}$, transition matrices $\{\Pimat^{(l)} \}_{l=1}^{L}$, and latent states $\{ \sv_{t,n}^{(l)} \}_{t=1,l=1,n=1}^{T,L,N}$.
While it is difficult to infer the introduced model for the coupling of $\{ \sv_{t,n}^{(l)} \}_{t=1,l=1,n=1}^{T,L,N}$ with $\{ \Phimat^{(l)} \}_{l=1}^{L}$ and $\{\Pimat^{(l)} \}_{l=1}^{L}$, the latent variables of the proposed model can be trained with a backward-upward--forward-downward (BUFD) Gibbs sampler described in Guo et al.~\cite{guo2018deep}, based on a variety of variable augmentation techniques.
However, the Gibbs sampler needs to process all HRRP data samples in each iteration and hence has limited scalability. For scalable inference, we adopt the topic-layer-adaptive stochastic gradient Riemannian (TLASGR) MCMC algorithm \cite{cong2017deep,Zhang2018WHAI}, which is proposed to update simplex-constrained global parameters~\cite{cong2017fast} in a mini-batch learning setting. Relying on the Fisher information matrix (FIM)~\cite{girolami2011riemann} to automatically adjust relative
learning rates for different parameters across all layers, TLASGR-MCMC has proven its improved
sampling efficiency. Given the latent states, we first sample augmented latent counts, then use TLASGR-MCMC to sample generative model parameters $\{ \Phimat^{(l)} \}_{l=1}^{L}$ and $\{\Pimat^{(l)} \}_{l=1}^{L}$.
To be more specific, we can sample $\piv_k^{(l)}$, the $k$th column of the transition matrix $\Pimat^{(l)}$, as
\begin{align}\label{TLASGR update_Pi}
\resizebox{.13\hsize}{!} {$ \left( {\piv_k^{(l)}} \right)_{i + 1}$} \! &= \resizebox{.8\hsize}{!} {$ \! \bigg[ \! \left( {\piv_k^{(l)}} \! \right)_i \! + \! \frac{\varepsilon _i}{M_k^{(l)}} \! \left[ \left(\rho \tilde \zv_{:k\cdotv}^{(l)} \! + \! \nuv^{(l)}\right) \! - \! \left(\rho \tilde z_{\cdotv k\cdotv}^{(l)} \! + \! \nu_{\cdotv}^{(l)} \! \right) \! \left( {\piv_k^{(l)}} \! \right)_i \right] $}\nonumber \\
\! &+ \! \resizebox{.7\hsize}{!} {$ \mathcal{N} \!\left( 0, \!\frac{2 \varepsilon_i}{M_k^{(l)}} \!\left[ \mbox{diag}(\piv_k^{(l)})_i \!-\! (\piv_k^{(l)})_i (\piv_k^{(l)})_i^T \!\right] \!\right) \! \bigg]_\angle $},
\end{align}
where {${\left[ . \right]_\angle }$ denotes the simplex constraint that $\piv_{k_1,k}^{(l)}\geqslant0$ and $ \sum_{k_1 = 1}^{K_l} \piv_{k_1,k}^{(l)}=1$}, $M_k^{(l)}$ is calculated using the estimated FIM, $\varepsilon_i$ denotes the learning rate at the $i$th iteration, both ${\tilde \zv_{:k\cdotv}^{\left( l \right)}}$ and ${\tilde z_{\cdotv k\cdotv}^{\left( l \right)}}$ come from the augmented latent counts $\Zmat^{(l)}$, and ${\nuv^{\left( l \right)}}$ denotes the prior of ${\piv_k^{\left( l \right)}}$.
More details of TLASGR-MCMC can be found in Cong et al.~\cite{cong2017deep} and Guo et al.~\cite{guo2018deep}.

Despite the attractive properties, both the proposed Gibbs sampler and TLASGR-MCMC usually rely on an iterative procedure to learn
the temporal latent states of a new HRRP sample at the testing stage, which hinders real-time processing of the HRRP based RATR.
To allow fast out-of-sample prediction, we further build an inference network to learn the latent states, as described below.

\textbf{Recurrent variational inference model:}
Our inference model is motivated by variational auto-encoders (VAEs)~\cite{kingma2014autoencoding}. As an example of a directed graphical model, the joint distribution over the observed variables $\xv$ and latent variables $\zv$ can be defined as
$p(\xv,\zv) = p(\xv \given\zv)p(\zv)$, where $p(\zv)$ is the prior placed on the latent variables. To admit efficient inference, VAEs approximate $p(\zv \given \xv)$ with a variational family of distributions $q(\zv \given \xv)$,
which adopts an inference network to map the observations directly to their latent space by optimizing the evidence lower bound (ELBO)~\cite{chung2015a}\cite{Jordan1999An}\cite{roberts18a}, expressed as
\begin{equation}\label{ELBO_ori}
L = \mathbb{E}_{q(\zv\given \xv)} [\ln p(\xv \given \zv)] - \mathbb{E}_{q(\zv \given \xv)} [ \ln [q(\zv \given \xv)/p(\zv)]].
\end{equation}
The approximate posterior $q(\zv\given\xv)$ is often taken to be a Gaussian distribution as $N(\uv,diag(\sigmav))$, where the mean $\uv$ and standard deviation $\sigmav$ are the output of highly non-linear functions of the input $\xv$.
Given the Gaussian variational posterior, the second term of the ELBO in \eqref{ELBO_ori} is analytic.
Using the reparameterization trick~\cite{kingma2014autoencoding}, VAEs sample $\zv$ with $\zv=\uv + \sigmav \odot \epsilonv$, where $\epsilonv$ is a vector of standard Gaussian variables. Therefore, the gradient of the first term of the ELBO with respect to the parameters of the inference network can be constructed as
\begin{gather}\label{derivative}
\small
 \nabla E_{q(\zv\given\xv)} [\ln p(\xv\given\zv) ] = E_{q(\epsilonv)} [ \nabla \ln p(\xv\given \zv=\uv + \sigmav \odot \epsilonv ) ],
 \end{gather}
whose estimation via Monte Carlo integration has low variance. The parameters of the inference model can be typically optimized via SGD.

VAEs provide an effective modeling paradigm for complex data distributions.
However, their success so far is mostly restricted to non-sequential data with Gaussian distributed latent variables and does not generalize well to model non-negative and sequential HRRP representations.
In this section, we propose a recurrent variational inference method to efficiently produce the multilayer temporal HRRP representations with \eqref{DPGDS_hidden} and \eqref{DPGDS_sample_PRG} or \eqref{Poisson_observed} as the generative model.
Given the global parameters $\{ \Pimat^{(l)}, \Phimat^{(l)}\}_{l=1}^{L}$, the task here is to infer the
latent states $\{ \sv_{t,n}^{(l)} \}_{t=1,l=1,n=1}^{T,L,N}$ via an inference network.
We first introduce a fully factorized distribution as
\begin{align}\label{posterior}
q\left(\left\{ \sv_{t,n}^{(l)} \right\}_{t=1,l=1,n=1}^{T,L,N}\right)= \prod\limits_{n = 1}^N {\prod\limits_{l = 1}^L {\prod\limits_{t = 1}^T {q\left( \sv_{t,n}^{(l)} \right) } } } .
 \end{align}
With \eqref{probability} and \eqref{posterior}, the objective function becomes
\begin{small}
\begin{align}\label{elbo}
 L\left( \left\{ q\left( {\sv_{t,n}^{(l)}}\right) \right \} _{l=1,t=1,n=1}^{L,T,N} ; \xv_{1:N}, \left\{ \Pimat ^{(l)}, \Phimat ^{(l)}\right\} _{l=1}^{L}\right) \notag \\
 = \sum\limits_{n = 1}^N {\sum\limits_{t = 1}^T {{\mathbb{E}_{q\left( {\sv_{t,n}^{(1)}} \right)}}\left[ {\ln p\left( {\xv_{t,n}\given{\Phimat ^{(1)}},\sv_{t,n}^{(1)}} \right)} \right]} } \\
 \!- \!\sum\limits_{n = 1}^N \! {\sum\limits_{l = 1}^L \! {\sum\limits_{t = 1}^T {{\mathbb{E}_{q\left( {\sv_{t,n}^{(l)}} \right)}} \! \left[
 { \ln \frac{\!{q \! \left( {\sv_{t,n}^{(l)}} \! \right)\!} }{ p\left( {\sv_{t,n}^{(l)}\given \av_{t,n}^{\left( l \right)},1/{b^{\left( l \right)}}} \right)}}\right]} } }\notag ,
\end{align}
\end{small}
where $\av_{t,n}^{\left( l \right)}:=b^{(l)}\left({{\Phimat ^{(l + 1)}}\sv_{t,n}^{(l + 1)} + {\Pimat ^{(l)}}\sv_{t{\rm{ - }}1,n}^{(l)}}\right)$ denotes the shape parameter of $\sv_{t,n}^{(l)}$.
Note under the BUFD Gibbs sampler~\cite{guo2018deep}, the conditional posterior of $\sv_{t,n}^{(l)}$ given augmented latent counts is gamma distributed, and hence
it might be more appropriate to use the gamma rather than Gaussian based distributions to construct
 the variational distribution $q\left( {\sv_{t,n}^{(l)}} \right)$.
However, the gamma random variable is not reparameterizable with respect to its shape parameter.
This motivates us to choose a surrogate distribution, which can be easily reparameterized, to approximate the gamma distribution.
The Weibull distribution is a desirable choice for this purpose, as its probability density function resembles that of the gamma distribution, and the second term in the ELBO shown in \eqref{elbo} becomes analytic if it is used to construct $q\left( {\sv_{t,n}^{(l)}} \right)$ \cite{Zhang2018WHAI}.

We may directly follow Zhang~et~al.~\cite{Zhang2018WHAI} to construct a Weibull distribution based inference network as
\begin{gather}\label{weibu}
q\left( {\sv_{t,n}^{(l)}} \right) \sim \mbox{Weibull}\left( {{\kv_{t,n}^{(l)}},{\lambdav _{t,n}^{(l)}}} \right),
\end{gather}
whose parameters $\kv_{t,n}^{(l)}$ and $\lambdav _{t,n}^{(l)}$ are both deterministically transformed from the hidden unit $\hv _{t,n}$ and specified as
\begin{align}
&\kv_{t,n}^{(l)} = \ln[1+\exp(\Wmat_{hk}^{(l)}\hv_{t,n}^{(l)}+ \bv_1^{(l)})] , \label{MLP1} \\
&\lambdav_{t,n}^{(l)} = \ln[1+\exp(\Wmat_{h \lambda}^{(l)}\hv_{t,n}^{(l)}+ \bv_2^{(l)})] , \label{MLP2}
\end{align}
where $\hv_{t,n}^{(l)}$ denotes the output of highly non-linear function of the observed $\xv_{t,n}$. However, this construction does not take into consideration the temporal information transmitted from the previous time step.
To exploit the temporal information, we propose a recurrent inference network which induces temporal
dependencies across time steps, as illustrated in Fig.~\ref{fig:deep_dynamical_model}~(c).
Therefore, similar to the RNN in \eqref{RNN}, we define $\hv_{t,n}^{(l)}$ as
\begin{equation}\label{RNN_update_h}
\small \hv_{t,n}^{(l)}=\left\{\begin{array}{ll}{\!\!\textrm{tanh}\left(\Wmat_{xh}^{(l)}\xv_{t,n}+\Wmat_{hh}^{(l)}\hv_{t-1,n}^{(l)}+ \bv_3^{(l)}\right)} ,& {\!\!\!\!\!\text { if } l=1 } ,\\
\!\!\textrm{tanh}\left(\Wmat_{xh}^{(l)}\hv_{t,n}^{(l-1)}+\Wmat_{hh}^{(l)}\hv_{t-1,n}^{(l)}+ \bv_3^{(l)}\right), & {\!\!\!\!\!\text { if } 1 < l\leq L} ,\\
\end{array}\right.
\end{equation}
where at layer $l$, $\Wmat_{xh}^{(l)}\in \mathbb{R}^{K_{l} \times K_{l-1}}$ denotes the upward weight matrix, $\Wmat_{hh}^{(l)} \in \mathbb{R}^{K_{l} \times K_{l}}$ the forward weight matrix connecting the hidden states, and $\bv_3^{(l)}\in \mathbb{R}^{K_{l} }$ the bias vector.
Therefore, the variational parameters of $\sv_{t,n}^{(l)}$ are both non-linearly transformed with neural networks from the hidden state $\hv_{t,n}^{(l-1)}$ of layer $l-1$ at time $t$ and the hidden state $\hv_{t-1,n}^{(l)}$ of layer $l$ at time $t-1$, which are helpful for the inference network to take into account the temporal structure of the observed data.

While the sequential latent states of the inference model are similar to that of an RNN, the internal transition structure of an RNN is in general entirely deterministic. By contrast, the proposed model introduces uncertainty into the latent space to help better model the variability observed in highly structured HRRP data.
One can sample the Weibull distributed latent state in \eqref{weibu} using the reparameterization trick as
\begin{align}\label{sample_theta}
\small
 \sv_{t,n}^{(l)} & \small = {\lambdav _{t,n}^{(l)}} \left(-\ln(1-{\epsilonv _{t,n}^{(l)}})\right) ^ {1/{\kv_{t,n}^{(l)}}},~
 {\epsilonv_{t,n}^{(l)}}\sim \mbox{Uniform}(0,1).
\end{align}

For standard VAE, the generative model parameters and the corresponding inference network parameters can be typically jointly
optimized via SGD, seeking to maximize the ELBO in \eqref{ELBO_ori}  with standard backpropagation technique. Instead of finding a point estimate of the global parameters of generative model like in VAEs, we adopt a hybrid MCMC/VAE inference algorithm by combining TLASGR-MCMC and the proposed recurrent variational inference network.
In specific, the generative model parameters $\{ \Pimat ^{(l)}, \Phimat ^{(l)}\} _{1,L}$ can be sampled with TLASGR-MCMC in \eqref{TLASGR update_Pi} and the neural network parameters $\Omegamat=\{\Wmat_{xh}^{(l)}, \Wmat_{hh}^{(l)},\Wmat_{hk}^{(l)}, \Wmat_{h \lambda}^{(l)},\bv_1^{(l)}, \bv_2^{(l)}, \bv_3^{(l)}\}_{1,L}$ can be updated via SGD by maximizing the ELBO in \eqref{elbo}. Applying the reparameterization trick of the Weibull distribution in \eqref{sample_theta}, the gradient of the ELBO with respect to $\Omegamat$ can be evaluated with low variance.
In practice, a single Monte Carlo sample from $q\left( \sv_{t,n}^{(l)} \right) $ is enough to obtain satisfactory performance.
The proposed inference network is depicted in Fig.~\ref{fig:deep_dynamical_model}~{(c)} and the training strategy is outlined in Algorithm \ref{Algorithm}.
For a new HRRP sample at the testing stage, given the generative model parameters and inferred network parameters $\Omegamat$, we can directly obtain the conditional posteriors of latent states using the inference network, without performing iterations.

\section{Supervised Variational Temporal Deep Generative Model}
The {rGBN} discussed above is an unsupervised model, which can infer the hierarchical latent states from HRRP samples under the condition of no class information.
Although the latent states can be used together with a downstream classifier to make label predictions, it is often beneficial to learn a joint model that considers both the HRRP samples and corresponding labels to discover more discriminative representations.
Therefore, we further develop a supervised rGBN (s-rGBN), providing multilayer latent representations that are good for both HRRP generation and classification.

Denote the $n$th sequential HRRP sample as a pair $\{\xv_n,y_n\}$, where $y_n \in \{1,2,...,C\}$ is the ground truth label of input $\xv_n$ and $C$ the total number of target classes. Assume that the HRRP label is generated from a categorical distribution {$p(y_{n}\given \xv_n, \Omegamat)\sim \mbox{Categorical}(p_{1n},...,p_{Cn})$}, written as
\begin{gather}\label{label}
p(y_n\given \xv_n, \Omegamat) = \prod \limits_{c = 1}^C {p_{cn}} ^{\textrm{I}\{y_n=c\}},
\end{gather}
where $p_{cn}$ is the probability of the current input $\xv_n$ classified to label $c$, $\sum_{c = 1}^C p_{cn}=1$, and $\textrm{I}\{y_n=c\}$ is an indicator function which is equal to one if $y_n=c$ and zero otherwise.

Since the introduced model is able to mine the deep hierarchical structure from the HRRP data, where the weight matrices at different layers reveal different levels of abstraction,
we combine the latent states across all hidden layers to define $p_{cn}$.
Note the sequential inputs from one HRRP sample share the same label, making its modeling different from a conventional sequence-to-sequence task addressed by RNNs.
We concatenate the latent state $\sv_{t,n}^{(l)}$ in \eqref{sample_theta} across all hidden layers and time steps to construct a latent feature vector of dimension $T\sum_{l = 1}^L {{K_l}} $, denoted as
\begin{gather}\label{feature}
 \Smat_n = \left[(\sv_{1,n}^{(1)},...,\sv_{T,n}^{(1)}),...,(\sv_{1,n}^{(L)},...,\sv_{T,n}^{(L)})\right].
\end{gather}
Given $\Smat_n$, the label probability vector $\pv_{n}=(p_{1n},...,p_{Cn})$ is calculated with the softmax function as 
\begin{gather}\label{softmax}
\small \pv_{n} =\left[ \frac{e^{\wv_1\Smat_n}}{\sum_{i=1}^{C} e^{\wv_i\Smat_n}},...,\frac{e^{\wv_c\Smat_n}}{\sum_{i=1}^{C} e^{\wv_i\Smat_n}},...,\frac{e^{\wv_C\Smat_n}}{\sum_{i=1}^{C} e^{\wv_i\Smat_n}}\right],
\end{gather}
where $\wv_c$ denotes the $c$th row of the learnable weight matrix $\Wmat_{sy}$. Since ${\Smat_n}$ is the concatenation of the latent states projected from $\xv_n$ using \eqref{weibu}, the label likelihood \eqref{label} can be rewritten as $\ p(y_n\given\Wmat_{sy},\Smat_n)$. The generative model for both the observed HRRP samples and labels can be displayed in Fig. \ref{fig:deep_dynamical_model}{(d)}.

Given the generative process, our proposed {s-rGBN} can be trained by maximizing the ELBO of the joint likelihood of the HRRP samples and labels, expressed as
 \begin{align}\label{loss_function}
\small
&\resizebox{0.7\hsize}{!}{$
L  =\sum\limits_{n = 1}^N {\sum\limits_{t = 1}^T {{E_{q\left( {\sv_{t,n}^{(1)}} \right)}}\left[ {\ln p\left( {\xv_{t,n}\given{\Phimat ^{(1)}},\sv_{t,n}^{(1)}} \right)}  \right]} }$}\notag\\
&~~~~\resizebox{0.95\hsize}{!}{$ -\sum\limits_{n = 1}^N \! {\sum\limits_{l = 1}^L \! {\sum\limits_{t = 1}^T {{E_{q\left( {\sv_{t,n}^{(l)}} \right)}} \! \left[
 { \ln \frac{\!{q \! \left( {\sv_{t,n}^{(l)}} \! \right)\!} }{ p\left( {\sv_{t,n}^{(l)}\given \av_{t,n}^{\left( l \right)},1/{b^{\left( l \right)}}} \right)} + \ln p(y_n\given\Wmat_{sy},\Smat_n)}\right]} } } $},
\end{align}
where the inference network for the latent states is the same as that of rGBN shown in Fig.~\ref{fig:deep_dynamical_model}~{(c)}. We also sample $\{ \Phimat ^{(l)},\Pimat ^{(l)} \}_{1,L}$ with TLASGR-MCMC, and update the neural network parameters $\Omegamat$ and classifier weight matrix $\Wmat_{sy}$ via SGD by maximizing the ELBO in \eqref{loss_function}, using the Adam optimizer~\cite{kingma2015adam}.
Moreover, based on the concatenation in \eqref{feature}, we notice that the supervised network enforces direct supervision for the latent states across all time steps and layers, improving the discriminativeness and robustness of the extracted latent states~\cite{lee2015deeply-supervised}.
At the test stage, given the learned inference network and classifier, s-rGBN can be very efficient on predicting the label of an HRRP sample $\xv_m$, expressed as
\begin{gather}\label{softmax}
 \setlength{\abovedisplayskip}{6pt}
 \hat{y}_m = \mathop {\arg \max } \limits_{c \in C} p\left(y_m=c\given\wv_c,\Smat_{m}\right),
\end{gather}
where $\Smat_{m}$ is the combination of $\sv_{t,m}^{(l)}$, which can be sampled with the reparameterization trick shown in \eqref{sample_theta}.
\begin{algorithm}[!t]
\scriptsize
\caption{Hybrid stochastic-gradient MCMC and recurrent variational inference for {rGBN}.}
\begin{algorithmic}
 \STATE Set mini-batch size $M$, the number of layer $L$, the width of layer $K_l$ and hyperparameters.
 \STATE Initialize inference model parameters $\Omegamat$, generative model parameters $ \{ \Pimat^{(l)}, \Phimat^{(l)}\}_{1,L}$.
 \FOR{$iter = 1,2, \cdots$ }
 \STATE
 Randomly select a mini-batch of $m$ HRRP samples to form
 a subset $\Xmat = \{ \xv_m \}_{1,M}$;\\
 Draw random noise $\{ {{\epsilonv _{t,m}^{(l)}}} \}_{t=1,m = 1,l=1}^{T,M,L}$ from uniform distribution \eqref{sample_theta};
 Sample latent states $\{\sv _{t,m}^{\left( l \right)}\}_{t=1,m = 1,l=1}^{T,M,L}$ from \eqref{sample_theta};\\
 Compute subgradient $ g = {\nabla _{\Omegamat}} L $ according to \eqref{elbo}, and update $\Omegamat$ using subgradient $g$;\\
 \FOR{$l = 1,2, \cdots,L$ and $ k =1, 2, \cdots, K_l $ }

 \STATE Update $M_{k}^{(l)}$ according to~\cite{cong2017deep}; then $\piv_{k}^{( l)}$ with \eqref{TLASGR update_Pi};\\
 Update $\phiv_{k}^{( l)}$ similar with \eqref{TLASGR update_Pi};\\
 \ENDFOR\\
 \ENDFOR\\
 Return global parameters $\{\Omegamat, \{ \Pimat^{(l)},\Phimat^{(l)} \}_{l=1}^{L}\}$.
\end{algorithmic}\label{Algorithm}
\end{algorithm}

\section{Experimental Results}
\subsection{Results with Synthetic data}
To illustrate the proposed model and compare it to existing methods, here we consider several three-dimensional synthetic datasets. Each toy data discussed below can be denoted as $\xv \in \mathbb{R}^{V \times T}$ with $V=3, T=100$.

$\bullet$ \textbf{Toy data 1:} To verify whether our proposed model can learn the transition matrix accurately, we generate the data with {rGBN-PRG}, where we
assume the transition matrix as $\Pimat = \begin{bmatrix} 0.65&0.20\\0.35&0.80\\ \end{bmatrix}$, initial latent state as $ \sv_{1}\! =\! \begin{bmatrix} 100 &0\\ \end{bmatrix} ^ \mathrm{T}$, and the Dirichlet distributed weight matrix as $\Phimat\in \mathbb{R}^{3 \times 2}$.
The latent states can be generated with $\sv_{t} \sim \mbox{Gam}\left( \Pimat \sv_{t-1}, 1\right)$, where $\sv_{t}\in \mathbb{R}^{2 \times 1}, T=100$. The non-negative observed data can be sampled from $\rv_{t}\sim \mbox{Pois}\left(\Phimat \sv_{t} \right),\xv_{t} \sim \mbox{Gam}\left(\rv_{t},1\right) $, where $\xv_{t}\in \mathbb{R}^{3 \times 1}$.

$\bullet$ \textbf{Toy data 2:}
{$x_{1,t}=t$, $x_{2,t}=2\exp(-t/15)+\exp(-((t-25)/10)^2)$, and $x_{3,t}= 5\sin(t^2) +6 $ for $t=1,\ldots,100$.}

$\bullet$ \textbf{Toy data 3:}
$x_{1,t} = t$, $x_{2,t} = 2\mbox{mod}(t,3)$, $x_{3,t}= 20\exp(-t/3)$ for $t=1,\ldots,50$, and
$x_{1,t} = 2t+30$, $x_{2,t} = 3\mbox{mod}(t,2)+5$, and $x_{3,t}= 30t\exp(-t)+10$ for $t=51,\ldots,100$, where $ \mbox{mod}(t,k)$ denotes the modulo operation that returns the remainder after division of $t$ by $k$.

Following previous work \cite{acharya2015nonparametric,gong2017deep,Adams2009}, we choose the particular expressions in Toy 2 and Toy 3 to check if our proposed model can capture such a complex temporal structure. We set the number of latent states as $K = 2$, and compare the proposed single layer {rGBN-PRG}, LDS, and HMM with both mean squared error (MSE) and prediction MSE (PMSE). MSE is measured between the estimated value and the ground truth for all the models at $t=1:T-1$, which is observed at the training stage; PMSE is measured between the predicted value and ground truth at the last time step, which is unobserved at the training stage.
Table~\ref{Tab:Results on Synthetic Data} illustrates the performance of different methods evaluated based on the simulated data.
It is clear that the proposed model provides satisfactory performance in both fitting and prediction for all datasets, which shows the benefits of capturing complex temporal structure in the latent space. For Toy 1, both {rGBN} and HMM are capable of discovering a transition matrix, $e.g.$, $\hat \Pimat=\begin{bmatrix} 0.712&0.164\\0.288&0.836\\ \end{bmatrix}$ in {rGBN} and
$\hat \Pimat=\begin{bmatrix} 0.924&0.018\\0.076&0.982\\ \end{bmatrix}$ in HMM, so the estimated transition matrix of the proposed model is closer to the ground truth than that of HMM.
For Toy 2, while LDS obtains lower MSE than the proposed model does, its PMSE is much worse, suggesting LDS is prone to overfitting on the training data. For Toy 3, we can find that features of dimensions 2 and 3 have very complex temporal relationships. HMM performs worse in Toy 3, possibly because of the complicated structure in Toy 3 makes HMM difficult to model.

\linespread{1.15}
\begin{table}[!htbp]
\vspace{-0.3cm}
\setlength{\abovecaptionskip}{0.cm}
\setlength{\belowcaptionskip}{-2.cm}
\centering
\caption{Results on Synthetic Data.}
\scalebox{0.85}{\begin{tabular}{ccccc}
\toprule[0.8pt]
Data& Measure & Our model & HMM & LDS \\
\hline
\multirow{2}*{\textbf{Toy1}} & MSE &\textbf{15.06} &20.08 &21.48 \\
& PMSE& \textbf{4.47} &9.83 &17.65\\
\hline
\multirow{2}*{\textbf{Toy2}} & MSE &2.11 & 27.59 & \textbf{1.21} \\
& PMSE& \textbf{2.47} &85.72 & 7.08\\
\hline
\multirow{2}*{\textbf{Toy3}} & MSE &\textbf{2.11} &53.98 &2.28 \\
& PMSE& \textbf{2.51} &250.69 &3.92\\
\bottomrule[0.8pt]
\end{tabular}\label{Tab:Results on Synthetic Data}}
\end{table}

\vspace{-0.5cm}
\subsection{{Measured HRRP data}}
A widely used HRRP dataset \cite{feng2017radar,du2011bayesian,Du2012Radar,XuBin}, consisting of the measurements from three real airplanes, is adopted to evaluate the performance of the proposed method. Yak-42 is a large and medium-sized jet aircraft, Cessna Citation is a small-sized jet aircraft, and An-26 is a medium-sized propeller aircraft.
The detailed parameters of the targets and measurement radar are presented in Table \ref{Tab:Parameters of radar and planes}.
We display the projections of the target trajectories onto the ground plane in Fig. \ref{fig:Projections}, where the measured data can be segmented into several parts.  Following the experiment settings in previous studies \cite{feng2017radar,Du2012Radar,XuBin,duAGC}, the measured HRRP data used in our experiments has been divided into training and testing data. There are two preconditions for selecting the training and test data. For one thing, the training data should contain almost all of the target-aspect angles of the test data. For another, the elevation angles of the training data should be different from those of the test data to verify the generalization ability of the proposed model. Therefore, we take the second and fifth segments of Yak-42, the sixth and seventh segments of Cessna Citation, and the fifth and sixth segments of An-26 for training, and take the other segments for testing. There are in total 7800 HRRP samples for training, corresponding to 2600 profiles for each of the three classes, and 5200 HRRP samples for testing. The number of target classes is $C=3$.

\vspace{-0.1cm}
\subsection{Set up}
For the networks used in this paper, the elements of all weight matrices are initialized with Gaussian distributions whose standard deviations are set to $0.1$, and all bias terms are initialized to $0$.
For the proposed models, we set the mini-batch size $M$ as $100$.
The Adam optimizer \cite{kingma2015adam} with learning rate ${10}^{-4}$ is utilized for optimization. Here we only choose a linear SVM (LSVM)~\cite{libsvm} for classifying the representations extracted with unsupervised models, in order to highlight the discriminability of the learned features. Real 256-dimensional HRRP vectors in \eqref{real_HRRP} are utilized to train the non-dynamical models. We set $[K_{1},K_{2},K_{3}] = [30, 20, 10]$ for the proposed methods.
All experiments are implemented in {non-optimized software written in Python, on a Pentium PC with 3.7-GHz CPU and 64 GB RAM.}
Unless specified otherwise, this setting will be adopted across all experiments.
For Bayesian generative model, to ensure the corresponding random variables drawn from non-informative priors and all the posteriors leaned from data, we directly set $\{\eta^{(l)}=0.1,\nu^{(l)}=0.1,b^{(l)}=1,c=200\}$ without tuning exhaustively. Besides, we set scaling parameter $\mu=100$, and manipulate the training and test data by a window function with window size $V=30$ and overlap $O=15$, leaving $T=16$ for each HRRP sequence, since they have the corresponding reasonable range as a priori knowledge. Below, instead of exhaustively optimizing these parameters, we demonstrate the influence of each parameter on model performance.

\linespread{1.3}
\begin{table}
\setlength{\abovecaptionskip}{0.cm}
\setlength{\belowcaptionskip}{-2.cm}
\centering
\caption{Parameters of radar and planes in the ISAR experiment.}
\scalebox{0.8}{\begin{tabular}{cccc}
\toprule[0.8pt]
\multirow{2}*{\textbf{Radar parameters}} & \textbf{Center Frequency} & \mc{2}{c}{5520MHZ}\\
&\textbf{bandwidth} & \mc{2}{c}{400MHZ}\\
\midrule[0.8pt]
\textbf{Planes}&\textbf{Length(m)}&\textbf{Width(m)}&\textbf{Height(m)}\\
\hline
{Yak-42} & 36.38 & 34.88 & 9.83\\
\hline
Cessna Citation &14.40 &15.90 &4.57\\
\hline
An-26& 23.80 &29.20& 8.58\\
\bottomrule[0.8pt]
\end{tabular}\label{Tab:Parameters of radar and planes}}
\end{table}

\vspace{-0.2cm}
\begin{figure}
\setlength{\abovecaptionskip}{-0.3cm}
\setlength{\belowcaptionskip}{-1cm}
\begin{center}
\centering
\includegraphics[height=3cm,width=8.8cm]{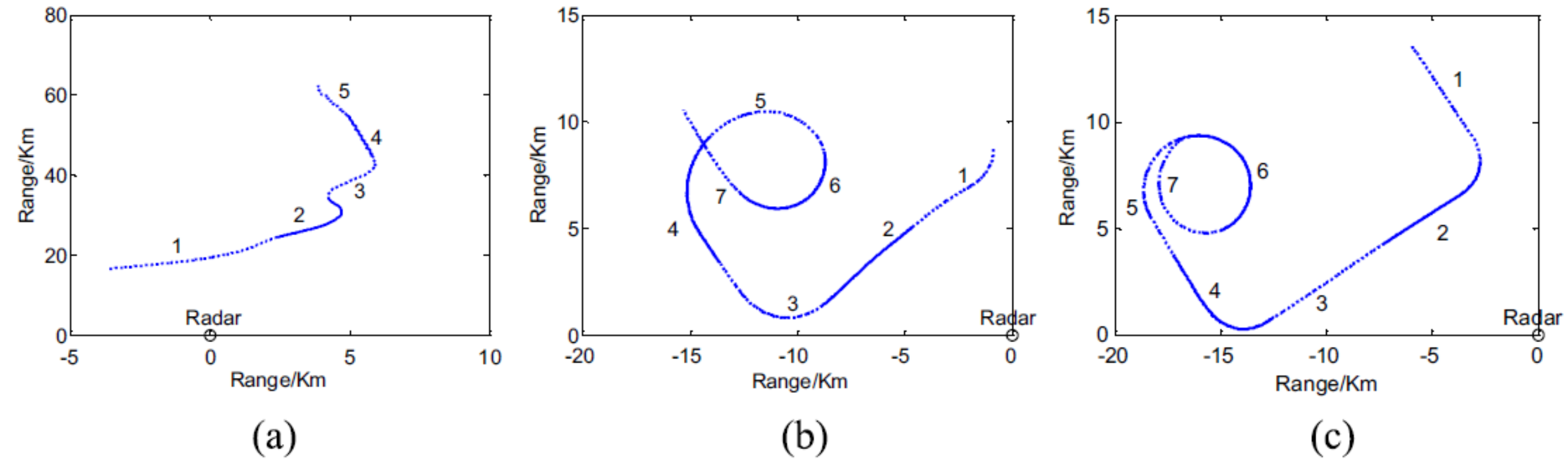}
\caption{Projections of the target trajectories onto the ground plane for (a) {Yak-42}, (b) Cessna Citation, and (c) An-26.}
\label{fig:Projections}
\end{center}
\end{figure}

\subsection{Influence of Model Parameters}
As shown in previous studies~\cite{feng2017radar,XuBin}, it is important to analyze how the recognition performance is influenced by the model parameters, which for the proposed models include the scaling parameter $\mu$, window size $V$, and overlap $O$.

\begin{figure}[!htbt]
\vspace{-0.3cm}
\begin{center}
\centering
\includegraphics[height=3.8cm,width=9cm]{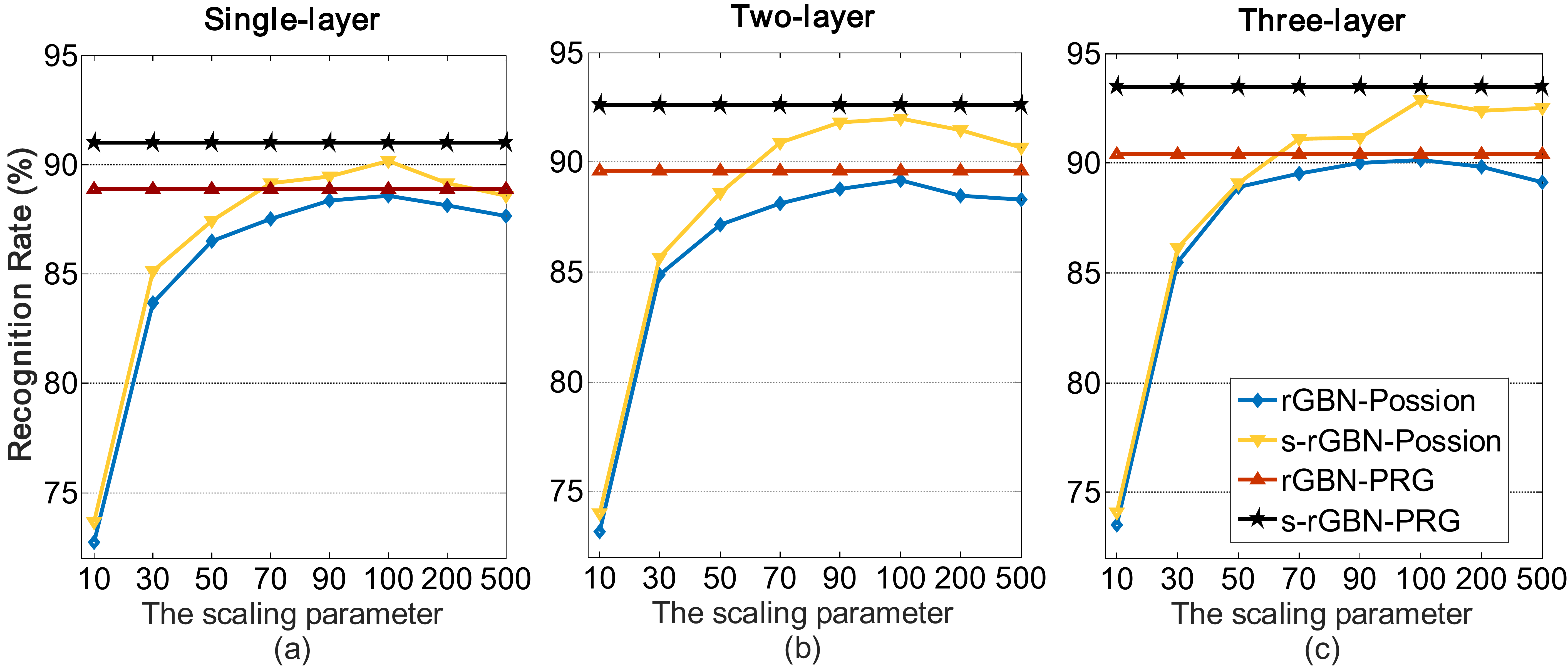}
\caption{Shown in (a)-(c) are the recognition performances for single-layer, two-layer, three-layer, respectively, as a function of the scaling parameter $\mu$, where the horizontal lines present the recognition performance of the PRG link, and the curves indicate the performances of the Poisson link for various values of $\mu$.}
\label{fig:different_scalar}
\end{center}
\end{figure}

\textbf{Scaling parameter $\mu$:}
As discussed in section \ref{my_model}, the sequential HRRP sample can be modeled with the PRG distribution in \eqref{DPGDS_sample_PRG}, where we refer our models as rGBN-PRG and s-rGBN-PRG. Considering the PRG link may be time-consuming due to its iterative procedure to infer latent counts, we discretize the HRRP sequences to produce counts ($i.e.$, non-negative integers) with the scaling parameter $\mu$. The input counts can be modeled with the Poisson distribution in \eqref{Poisson_observed}, where we refer our models as rGBN-Poisson and s-rGBN-Poisson.
Fig. \ref{fig:different_scalar} shows the recognition performances of our proposed models varying with $\mu$. Clearly, a large value of $\mu$ will lead to overfitting and a small one will drop too much detailed information of the HRRP data, resulting in underfitting. For our proposed models, although using the PRG link generally perform better than using the Poisson link, we can achieve a compromise between the performance and computation with the Poisson link.

\begin{figure}[!htbt]
\begin{center}
\centering
\includegraphics[height=3.7cm,width=9cm]{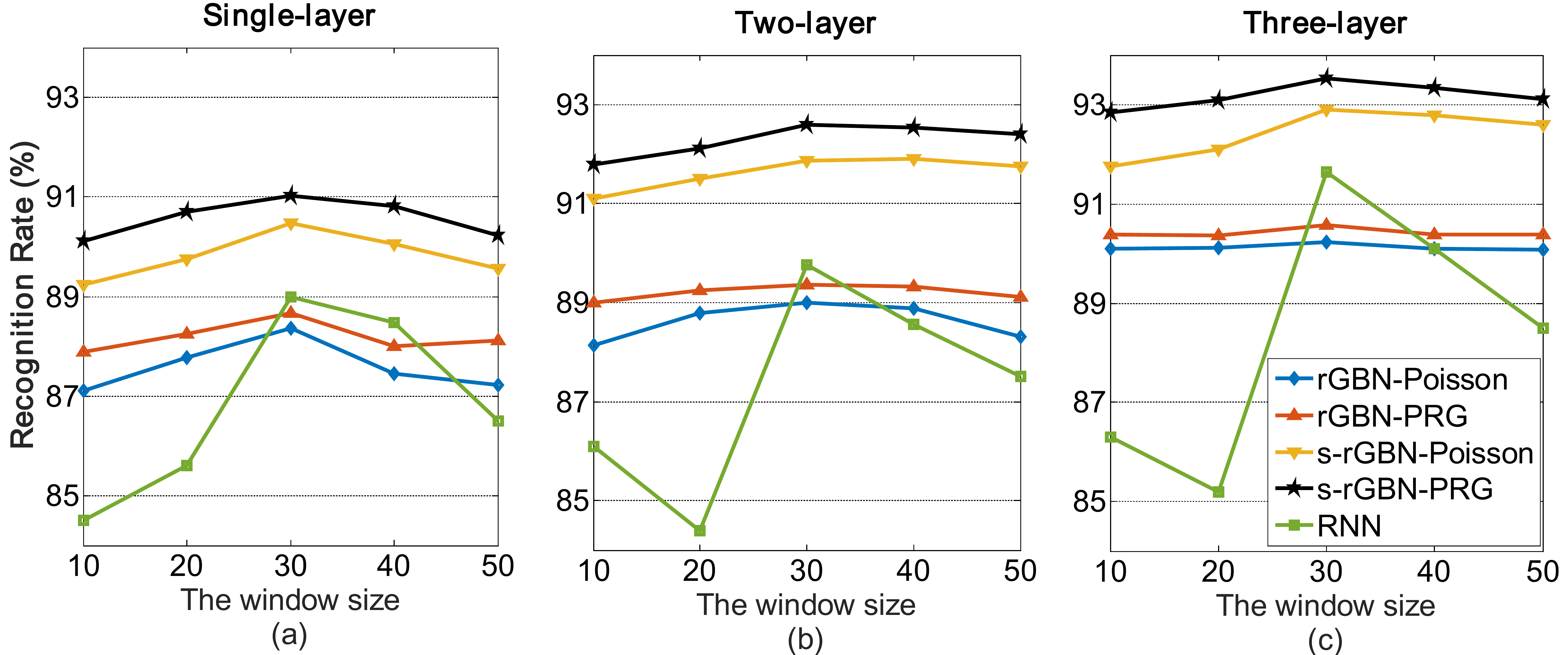}
\caption{Shown in (a)-(c) are the recognition performance for single-layer, two-layer, three-layer, respectively, as a function of the window size $V$ for the dynamical models.}
\label{fig:different_v}
\end{center}
\end{figure}

\textbf{Window size $V$:}
Fig. \ref{fig:different_v} shows the variation of the classification accuracies of dynamical models with the value of window size $V\in [10,50]$, where the overlap length is fixed as $O = V /2$.
As $V$ decreases, we get the sequence with a higher computational and memory burden for all temporal models, thus increasing the inference difficulty and degrading their performance.
On the contrary, a large $V$ allows more information of the target to be contained in each subsequence $\xv_{t,n}$, but a too-large one may result in the loss of sequential information.
Fig. \ref{fig:different_v} shows that compared with RNN, our proposed models, which provide a stochastic generalization of the deterministic RNN by adding uncertainty into the latent space via a deep generative model, are more robust to the window length.

\textbf{Overlap $O$:}
After fixing the window size as $V=30$, we compare the performance of different dynamical methods as a function of the overlap length $O$, which varies from 5 to 25 range cells.
Because the redundancy of the segments is determined by the overlap length across the windows, a lower correlation between adjacent inputs is encouraged as parameter $O$ decreases, which will weaken the sequential relationship between time steps. Instead, an improperly large one will raise the length of sequence fast, imposing higher computational and memory burden and reducing the performance. Besides, the proposed models are more robust to the overlap length compared with RNN.
\begin{figure}[!htbt]
\begin{center}
\centering
\includegraphics[height=3.7cm,width=9cm]{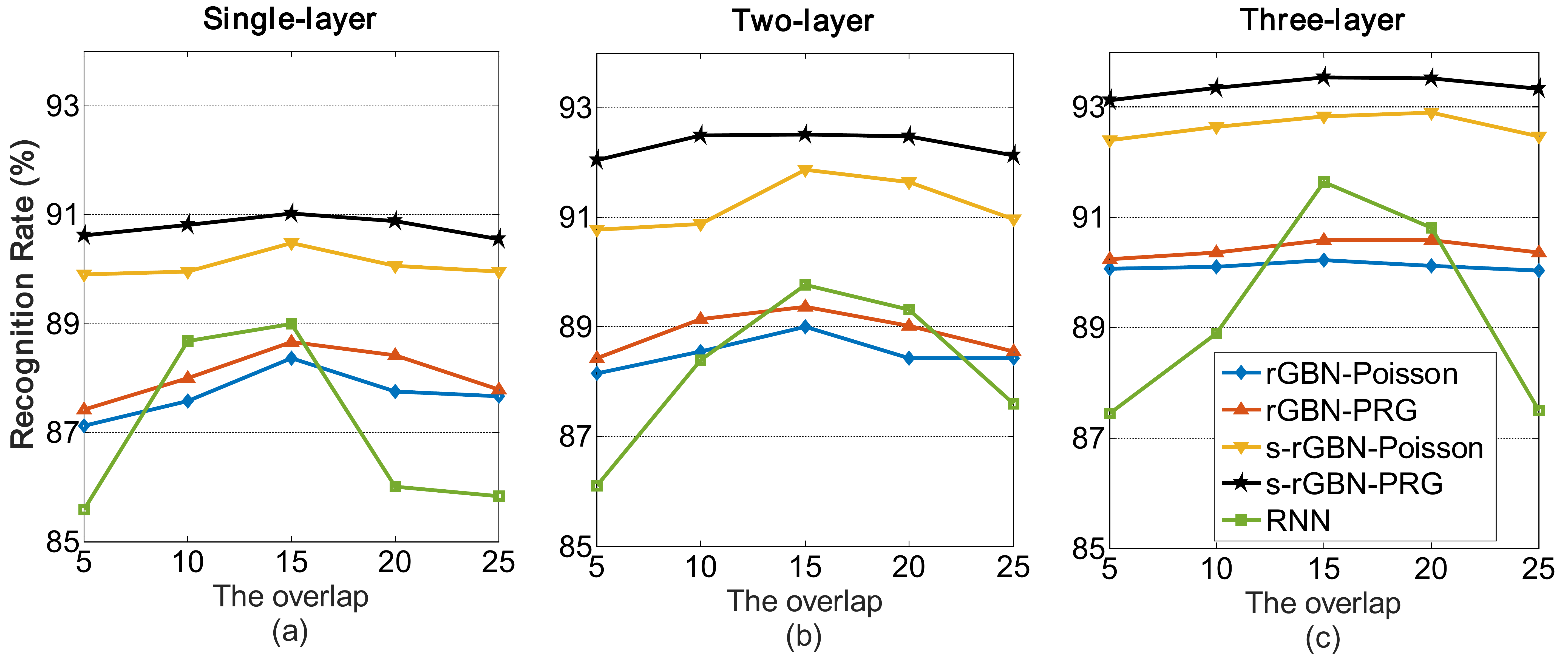}
\caption{{Shown in (a)-(c) are the recognition performance for single-layer, two-layer, three-layer}, respectively, as a function of the overlap $O$ for various dynamical models.}
\label{fig:different_overlap}
\end{center}
\end{figure}
\vspace{-0.6cm}
\subsection{Recognition Performance}
\vspace{-0.1cm}
\begin{table}[htbp]
\setlength{\abovecaptionskip}{-3pt}
\setlength{\belowcaptionskip}{-3pt}
 \centering
 \caption{Comparison of Recognition Accuracy}
 \label{tab:Margin_settings}
 \scalebox{0.8}{\begin{tabular}{ccc}
 \toprule
 \textbf{Non-dynamical Models} &\textbf{Size} &\textbf{Accuracy} \\
 \hline
 MCC \cite{du2005radar} &- & 59.00\\
 AGC \cite{duAGC} &- & 85.20\\
 LSVM \cite{libsvm} &- & 87.14\\
 LDA \cite{yu2001a} &2 & 82.16\\
 PCA \cite{Du2007Radar} &200 & 83.81\\
 VAE \cite{kingma2014autoencoding} &200 & 87.84\\
 \hline
 \mr{3}*{DBN \cite{hinton2006a}} &200 & 88.28\\
 &200-100 & 88.51\\
 &200-100-50 & 89.16\\
 \toprule
 \textbf{Dynamical Models} &\textbf{Size} &\textbf{Accuracy} \\
 \hline
 LDS \cite{WangLDS} &30 &87.65 \\
 HMM \cite{Pan2012Multi} &30 & 87.24\\
 {TARAN \cite{XuBin}} &30 &90.10\\
 \hline
 {TCNN \cite{wan2019convolutional}} & 1 $\times$ 9 (32-32-64) &92.57\\
 \hline
 \mr{3}*{RNN \cite{graves2013speech}} &30 & 88.99\\
 &30-20 & 89.77\\
 &30-20-10 & 91.64\\
 \hline
 \mr{3}*{rGBN-Poisson} &30 & 88.36\\
 &30-20 & 89.22\\
 &30-20-10 &90.23\\
 \hline
 \mr{3}*{rGBN-PRG} &30 &88.66 \\
 &30-20 &89.67 \\
 &30-20-10 &90.58\\
 \hline
 \mr{3}*{s-rGBN-Poisson} &30 &90.47 \\
 &30-20 &91.87 \\
 &30-20-10 &92.91\\
 \hline
 \mr{3}*{s-rGBN-PRG} &30 &91.02 \\
 &30-20 &92.52 \\
 &30-20-10 &\textbf{93.54} \\
 \bottomrule
 \end{tabular}\label{Tab:results}}
 \end{table}

Similar to Feng~et~al.~\cite{feng2017radar}, we evaluate the proposed models (rGBN and s-rGBN) against several commonly used recognition methods for HRRP, such as maximum correlation coefficient (MCC)~\cite{du2005radar}, adaptive Gaussian classifier (AGC)~\cite{duAGC}, and LSVM using the original HRRP dataset as input~\cite{libsvm}.
A variety of feature extraction methods, including linear discriminant analysis (LDA)~\cite{yu2001a}, PCA~\cite{Du2007Radar}, VAE~\cite{kingma2014autoencoding}, LDS~\cite{WangLDS}, and HMM~\cite{Pan2012Multi}, are also included for comparison.
To demonstrate the advantages of constructing deep dynamical systems, we further consider RNN~\cite{graves2013speech} and deep belief network (DBN)~\cite{hinton2006a}, where DBN can be viewed as a stack of restricted Boltzmann machines (RBMs) for modeling the binary hidden units in the lower layers. In addition to the basic RNN, we also compare the results of the proposed method with target-aware recurrent attentional network (TARAN) \cite{XuBin}. TARAN first utilizes RNN to explore the sequential relationship between the range cells within an HRRP sample, then employs an attention mechanism to weight up each hidden state and discover the target area. We further compare our
model with temporal one-dimensional convolution neural network (TCNN) \cite{wan2019convolutional}, where the convolution operation only takes place along the range dimension. Note TARAN \cite{XuBin} presents the recognition result of the time sequential HRRP only with one hidden layer (hidden dimension is $30$), achieving the performance of 90.10\%, and TCNN \cite{wan2019convolutional} achieves a recognition accuracy of 92.57\% with the structure of 32-32-64 (kernel size is $1 \times 9$ in each layer).
For dynamical models, we set the number of hidden states of LDS and HMM as $30$, and set that of the RNN the same as that of the proposed models.
The feature dimension of both PCA and VAE is set as 200, while that of DBN as $[K_{1},K_{2},K_{3}] = [200, 100, 50]$, which all belong to non-dynamical feature extraction models.
According to the literature~\cite{feng2017radar}, the hidden dimension of LDA can be set as $C-1=2$.
The extracted features from the training set with different unsupervised models are utilized to train the LSVM classifier, where the regularization parameter is five-fold cross-validated on the training set.
The softmax function shown in \eqref{softmax} is adopted to predict the labels of the testing samples for supervised models, including s-rGBN and RNN.
Summarized in Table \ref{Tab:results} are the recognition results of various methods on the HRRP dataset.

The shallow dynamical models including LDS, HMM, and TARAN, which take into account the temporal information in HRRP, already clearly outperform other traditional models for HRRP RATR including AGC, MCC, LSVM, LDA, and PCA. Besides, the deep models tend to have better performance on classifying HRRP samples than shallow architectures do, for providing more discriminative hierarchical non-linear features.
As for DBN, although it builds a deep generative model, we find that its performance is inferior to our proposed deep dynamical models, which is not surprising as the former ignores the temporal dependence in HRRP samples. Compared with TCNN, RNN, and TARAN, which build the deterministic mapping and find point estimates for the global parameters, s-rGBN-PRG achieves better accuracy, proving the efficiency of the probabilistic framework and the hybrid Bayesian inference algorithm.

Note that our proposed s-rGBN-PRG and s-rGBN-Poisson perform better than the unsupervised rGBN-PRG and rGBN-Poisson, which verifies that introducing the label information into proposed models certainly benefits the recognition performance.
Though our proposed unsupervised rGBN schemes underperform supervised models, such as RNN at layer 3 and TCNN, they still can learn hierarchical latent states from HRRP data and obtain acceptable recognition rates in the absence of label information.

\setlength{\abovetopsep}{0.5ex}
\setlength{\belowrulesep}{0pt}
\setlength{\aboverulesep}{0pt}
\linespread{1.3}
\begin{table*}[!htbp]
\centering
\caption{The confusion matrices of the proposed s-rGBN at different layers, with the network architecture set as 30-20-10.}
\scalebox{0.8}{\begin{tabular}{c|ccc|ccc|ccc}
\toprule[0.8pt]
\textbf{Methods} & \mc{3}{c|}{\textbf{ s-rGBN-layer1}} & \mc{3}{c|}{\textbf{ s-rGBN-layer2}} &\mc{3}{c}{\textbf{ s-rGBN-layer3}}\\
\cline{2-4}
\cline{5-7}
\cline{8-10}
 & An-26 & Cessna Citation &{Yak-42} & An-26& Cessna Citation & {Yak-42} & An-26 & Cessna Citation & {Yak-42}\\
\midrule[0.8pt]
An-26& 86.17 &8.63 & 5.20& 87.87 &7.08 &5.05 & 89.39 &6.30 &4.31 \\
\hline
Cessna Citation &5.30 &93.45 &1.25 &4.30 &94.40 & 1.30 & 4.05 &95.30 &0.65 \\
\hline
{Yak-42} &3.08 &2.17 & 94.75 & 1.84 &1.33 &96.83 & 1.57 & 1.18 &97.25 \\
\hline
Average Recognition Rates& \mc{3}{c|}{91.02} &\mc{3}{c|}{92.52} & \mc{3}{c}{\textbf{93.54}} \\
\bottomrule[0.8pt]
\end{tabular}\label{Tab:confusion matrices}}
\end{table*}
To investigate the performance of the methods on different targets, we list the confusion matrices and average recognition rates in Table \ref{Tab:confusion matrices}.
Each row of the confusion matrix should be the number of samples in a specific predicted class. For the sake of comparison, we set each row of the matrix as a predicted ratio rather than a number. Specifically, we set the confusion matrix of each method as $P \in \mathbb{R}_{+}^{C\times C}$, and $P_{c_1,c_2} = {N_{c_1,c_2}}/{N_{c_1}}$ , where $N_{c_1}$ denotes the number of HRRP samples for category $c_1$, $N_{c_1,c_2}$ denotes the the number of samples which are in an actual class $c_1$ but are classified into class $c_2$, and $\sum_{c_2=1}^C P_{c_1,c_2} = 1$. There is a clear trend that the classification performance for the HRRP samples based on s-rGBN is increasing with more layers, where the gains of the accuracy mainly come from the improvement at layer 2 for Yak-42 and Cessna Citation and that at layer 3 for An-26.
The An-26 aircraft is a propeller aircraft whose waveform has larger fluctuation, and hence its feature may be much more overdispersed in comparison to these of Cessna Citation and Yak-42. s-rGBN, whose learned neurons become more general with the increase of the layers, maybe more robust to the fluctuation in higher layers and hence learn discriminative features to improve the accuracy.

\subsection{Qualitative Analysis}
As discussed in Section III, in comparison to RNN that builds temporal non-linearity via ``black-box'' neural networks and other dynamical methods that only model single-layer temporal latent states, our proposed models can provide the desired interpretation, for being a deep Bayesian dynamical generative model.
In particular, in addition to quantitative evaluations, we further visualize the learned neurons, reconstruction examples, latent space interpolations on the HRRP test set, and learned latent states, at each layer.

\begin{figure}[!htbt]
\vspace{-0.3cm}
\setlength{\abovecaptionskip}{-0.3cm}
\setlength{\belowcaptionskip}{-1cm}
\begin{center}
\centering
\includegraphics[height=3cm,width=6.8cm]{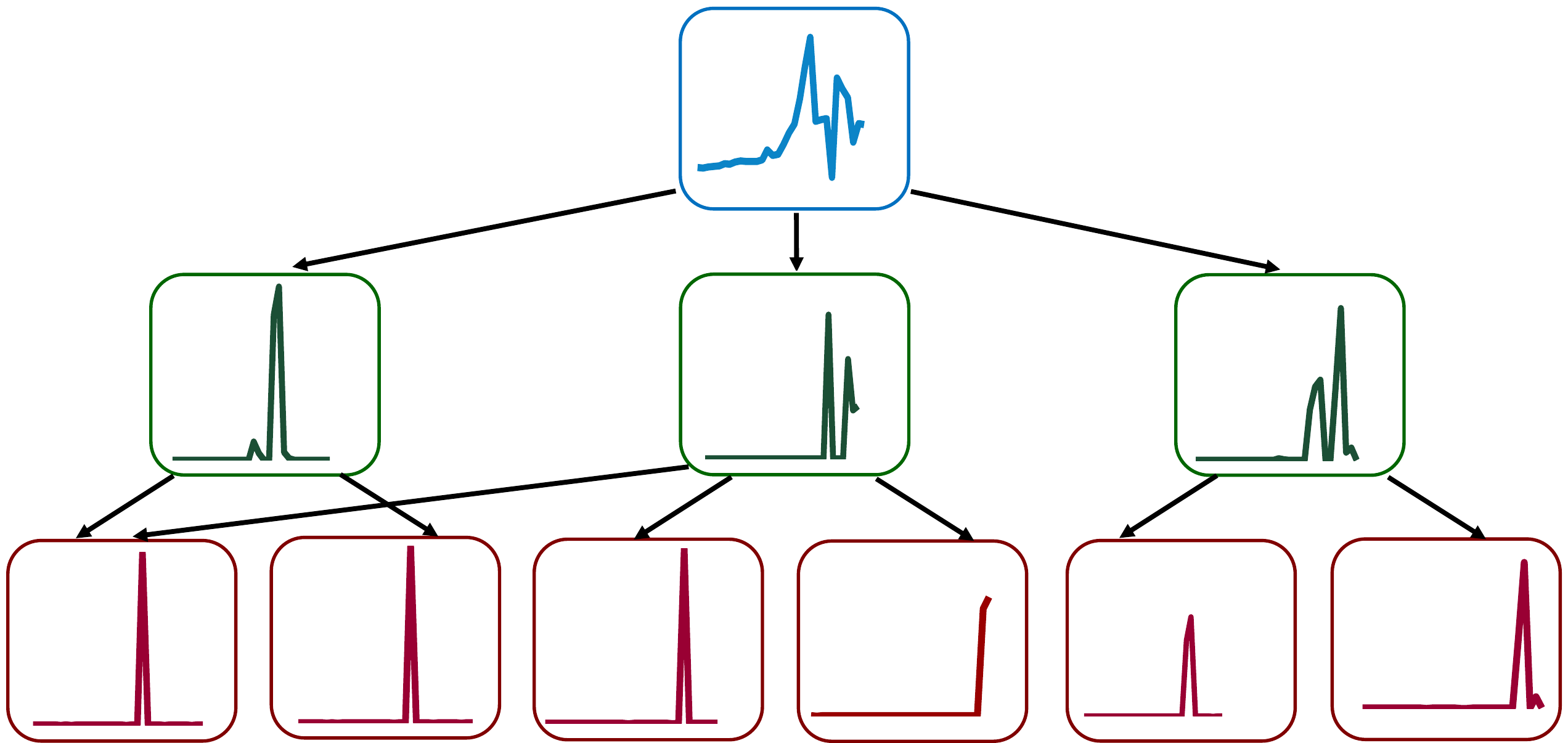}
\caption{Visualization and hierarchy of example neurons in different layers. Neurons at layers 3, 2, and 1 are shown in blue, green, and red boxes, respectively.}
\label{fig:dictionary}
\end{center}
\end{figure}

\textbf{The structured neurons:}
To visualize the multilayer s-rGBN, it is straightforward to project the neuron $\phiv_{k}^{(l)}$ of layer $l$ to the bottom data layer.
Specifically, we first choose a node at the top layer, with a large coefficient, then grow the tree downward to include any lower-layer neurons that are connected to the node with non-negligible weights. For $\phiv_{k}^{(l)}$, we plot all its terms whose values are larger than 1\% of the largest element of the corresponding $\phiv_{k}^{(l)}$.
As shown in Fig.~\ref{fig:dictionary}, we find that the neurons at layer 1 are fundamental, similar to the original echo per range cell in HRRP.
As layer $l$ increases, the neurons become increasingly more general or sparsely utilized, which consist of several echoes from different range cells in HRRP, covering longer-range temporal dependencies. To sum up, our proposed model can not only extract meaningful neurons at each layer but also capture the relationships between the neurons of different layers.

\begin{figure}[!htbt]
\vspace{-0.3cm}
\setlength{\abovecaptionskip}{-0.3cm}
\setlength{\belowcaptionskip}{-10cm}
\begin{center}
\centering
\includegraphics[height=6.9cm,width=8.3cm]{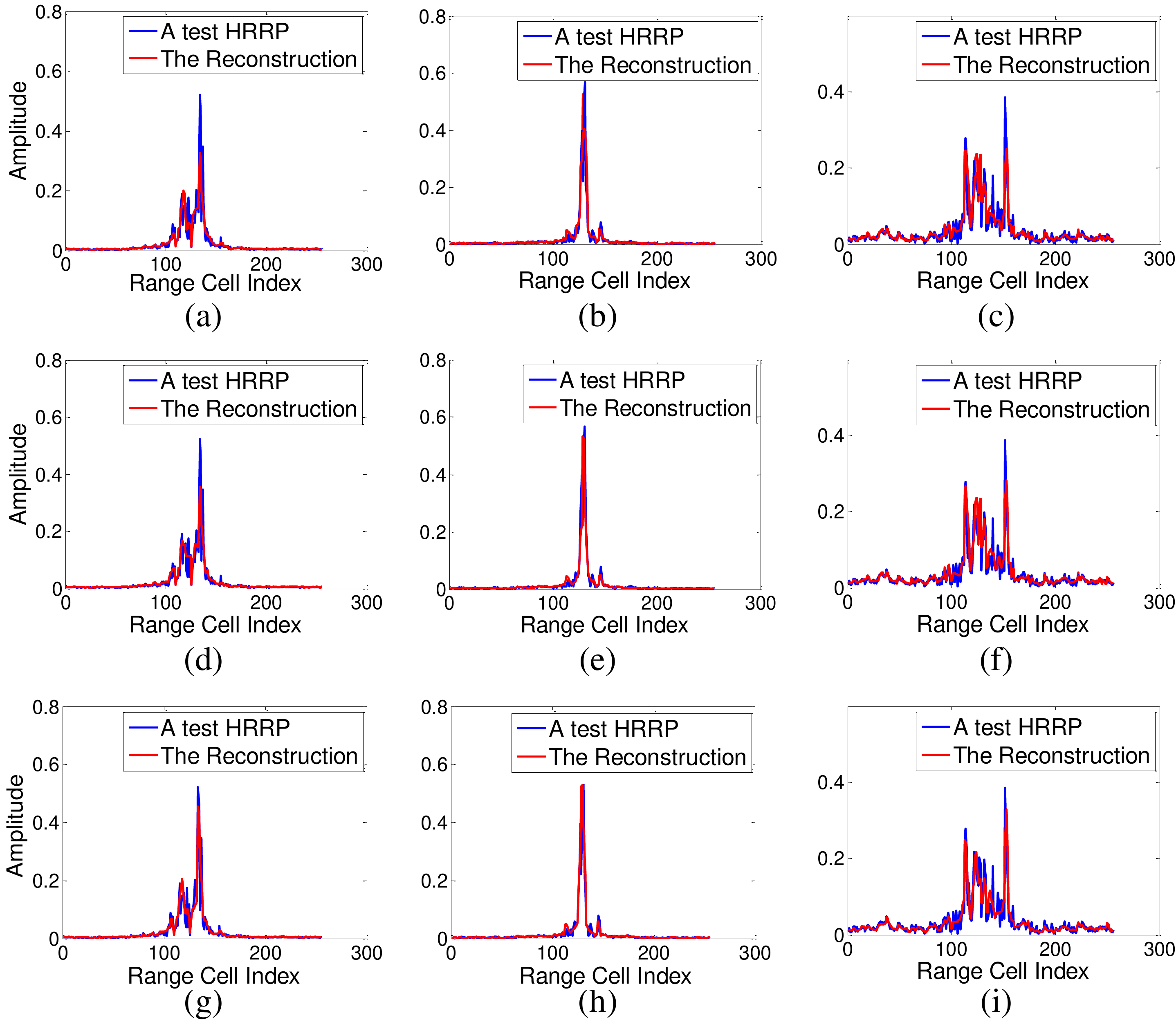}
\caption{The reconstruction performance of s-rGBN for three testing samples at different layers.
\textbf{\textrm{Columns 1-3}}: The reconstructing samples of An-26, Cessna Citation, and Yak-42, respectively;
\textbf{\textrm{Rows 1-3}}: The reconstructing samples based on s-rGBN layer-1, layer-2, and layer-3, respectively.
}
\label{fig:reconstruction_all}
\end{center}
\end{figure}

\textbf{Reconstruction:}
In Fig. \ref{fig:reconstruction_all}, we show the reconstruction examples for the three airplane targets at layers 1, 2, and 3.
To reconstruct the HRRP samples, {given the learned global parameters $\{\Omegamat, \{ \Phimat^{(l)},\Pimat^{(l)} \}_{l=1}^{L}\}$}, we need to find the conditional posterior of latent state $\sv_{t,n}^{(l)}$ , whose variational parameters can be directly transformed from the observed HRRP examples using the neural networks in \eqref{MLP1}, \eqref{MLP2}, and \eqref{RNN_update_h}.
Then we sample the latent states using the reparameterization trick in \eqref{sample_theta}, where a single Monte Carlo sample from $q\left( \sv_{t,n}^{(l)} \right) $ is enough to obtain satisfactory performance.
We find that our model can not only retain the main structural information of the original test HRRP samples but also reconstruct the details of the targets in each sequential HRRP.

\begin{figure}[!htbt]
\setlength{\abovecaptionskip}{-0.3cm}
\setlength{\belowcaptionskip}{-0.3cm}
\begin{center}
\centering
\includegraphics[height=4.3cm,width=7.7cm]{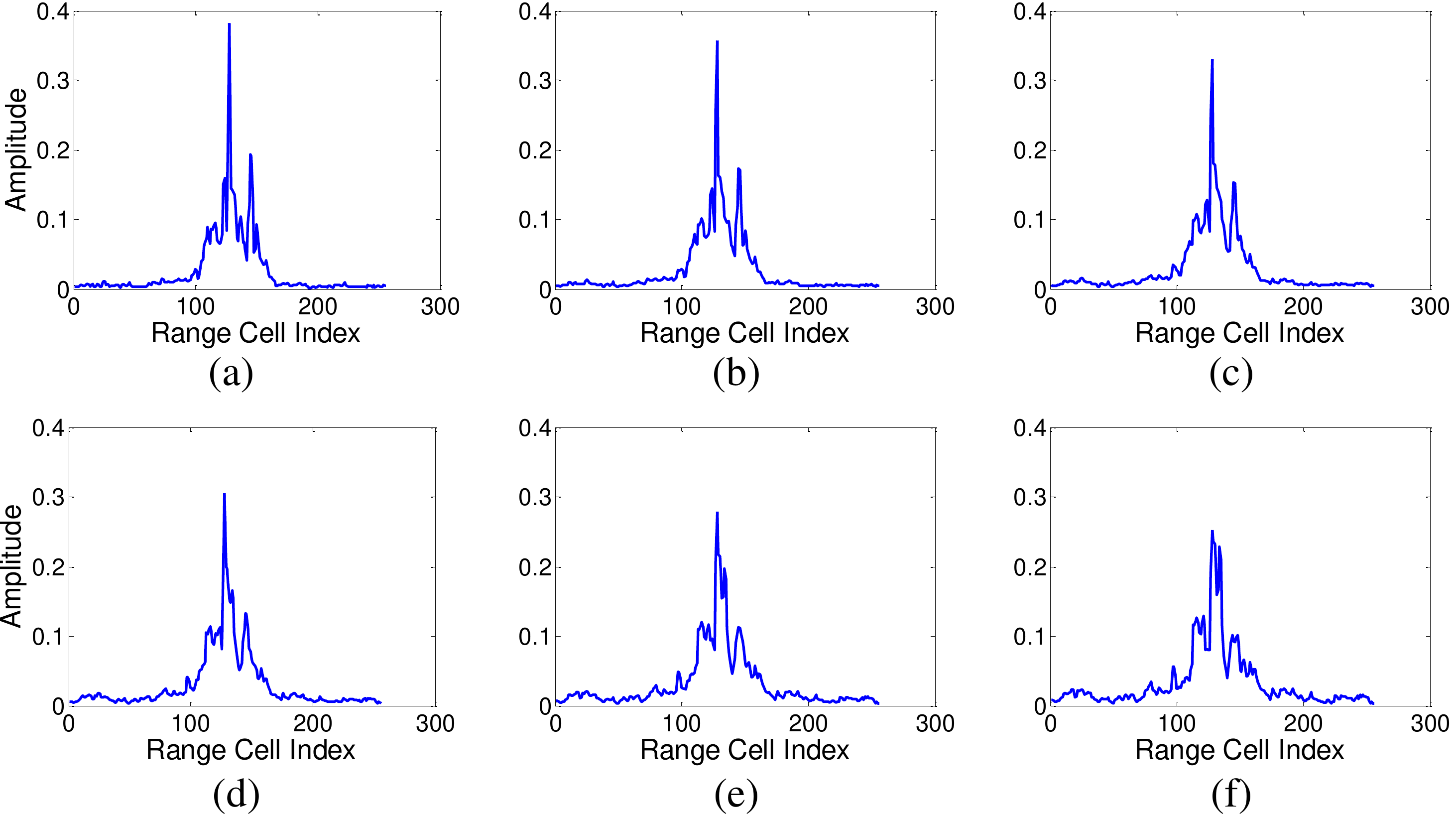}
\caption{Latent space interpolations on HRRP test set. (a) and (f) are the samples generated from $\sv_{1:T,1}^{(3)}$ and $\sv_{1:T,2}^{(3)}$, respectively, and the others are generated from the latent states interpolated linearly from $\sv_{1:T,1}^{(3)}$to $\sv_{1:T,2}^{(3)}$.}
\label{fig:latent}
\end{center}
\end{figure}

\textbf{Latent space interpolations:}
One might want to investigate whether s-rGBN has indeed {extracted} the abstract representations of HRRP data.
Following previous ideas \cite{Zhang2018WHAI,dumoulin2017adversarially}, we present the latent space interpolations on the HRRP test set examples.
Given two HRRP sequences $\xv_{1:T,1}$ and $\xv_{1:T,2}$, we project them into $\sv_{1:T,1}^{(1:3)}$ and $\sv_{1:T,2}^{(1:3)}$, with the 3-layer model learned before.
Using linear interpolation between $\sv_{1:T,1}^{(l)}$ and $\sv_{1:T,2}^{(l)}$ at layer $l$, we can produce new sequences by passing the intermediate points through the
dynamical generative model.
 In Fig. \ref{fig:latent}, we display the reconstruction results of $\sv_{1:T,1}^{(3)}$ and $\sv_{1:T,2}^{(3)}$ in (a) and (f), respectively, and
the generated sequences from the linearly interpolated $\sv$-values in (b)-(e).
The proposed model can generate realistic and interpretable HRRP sequences
for all interpolated $\sv$-values. In other words, the inferred latent space of the model is on a manifold, indicating our proposed model has learned a generalizable latent state representation instead of concentrating its probability mass around the HRRP training samples.

\begin{figure}[!htbt]
\setlength{\abovecaptionskip}{-0.3cm}
\setlength{\belowcaptionskip}{-0.3cm}
\begin{center}
\centering
\includegraphics[height=5.5cm,width=7.7cm]{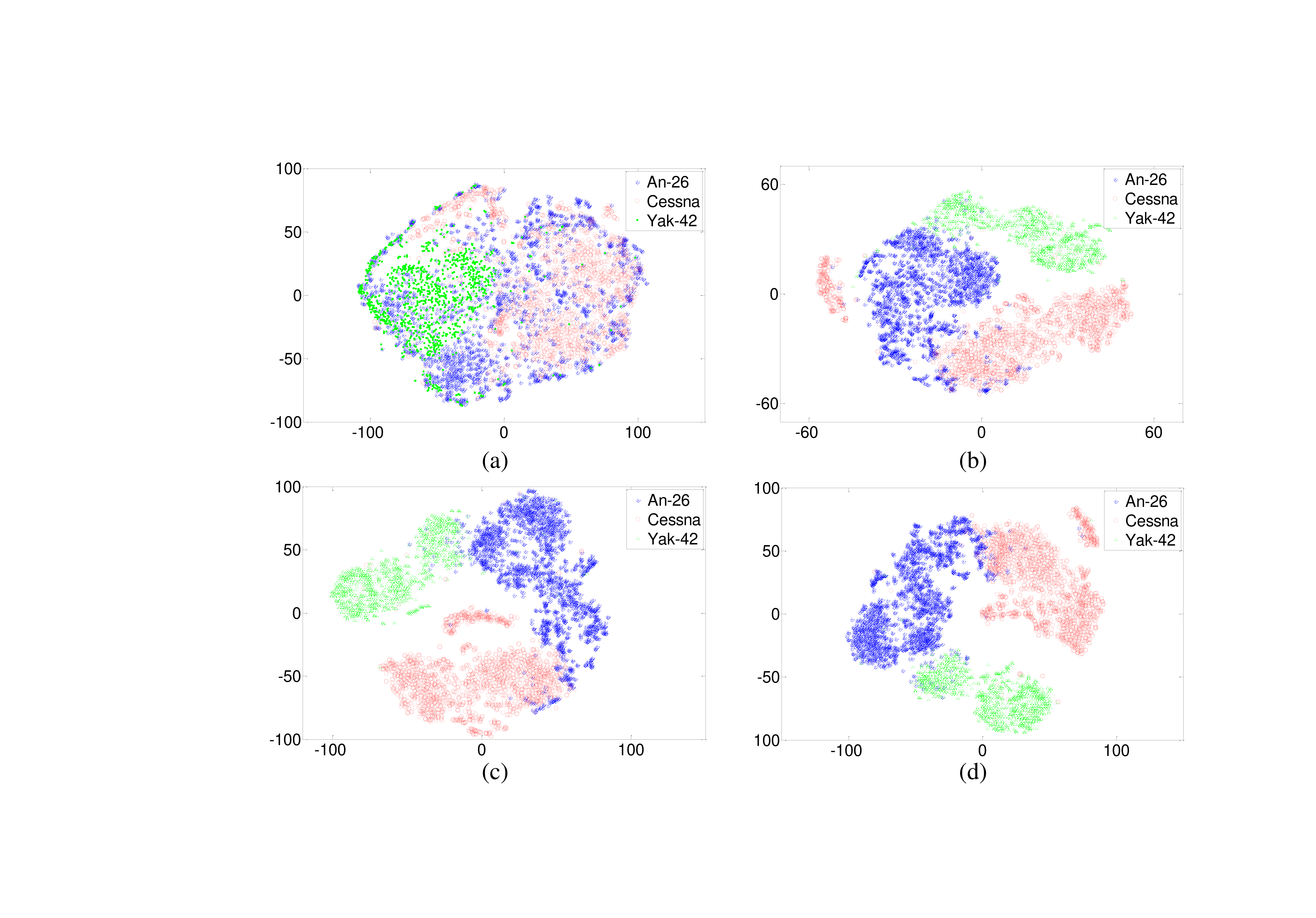}
\caption{Two-dimensional $t$-SNE projection of the test HRRP samples and their corresponding features at different layers.
Shown in (a) are the original test HRRP samples, and shown in (b)-(d) are learned features in rGBN, RNN, and s-rGBN, respectively.}
\label{fig:tsne}
\end{center}
\end{figure}

\textbf{Latent states:}
To compare the latent states learned from different methods, we visualize high-dimensional features by mapping them to the two-dimensional subspace with $t$-distributed stochastic neighbor embedding ($t$-SNE). It is a non-linear dimensionality reduction technique well-suited for embedding high-dimensional data into a low-dimensional space~\cite{dermaaten2008TSNE}. As shown in Fig. \ref{fig:tsne}, each dot represents an HRRP sample and each color-shape pair denotes a category.
We visually illustrate that the features learned by the proposed supervised method are more discriminative than those by the unsupervised method and RNN. The phenomenon proves the benefits of learning a supervised hierarchical probabilistic model that considers both target and label generation.

\vspace{-0.3cm}
\subsection{{Robustness to Training Data Size}}
\begin{figure}[!htbt]
\vspace{-0.2cm}
\setlength{\abovecaptionskip}{-0.3cm}
\setlength{\belowcaptionskip}{-7cm}
\begin{center}
\centering
\includegraphics[height=2.7cm,width=5.8cm]{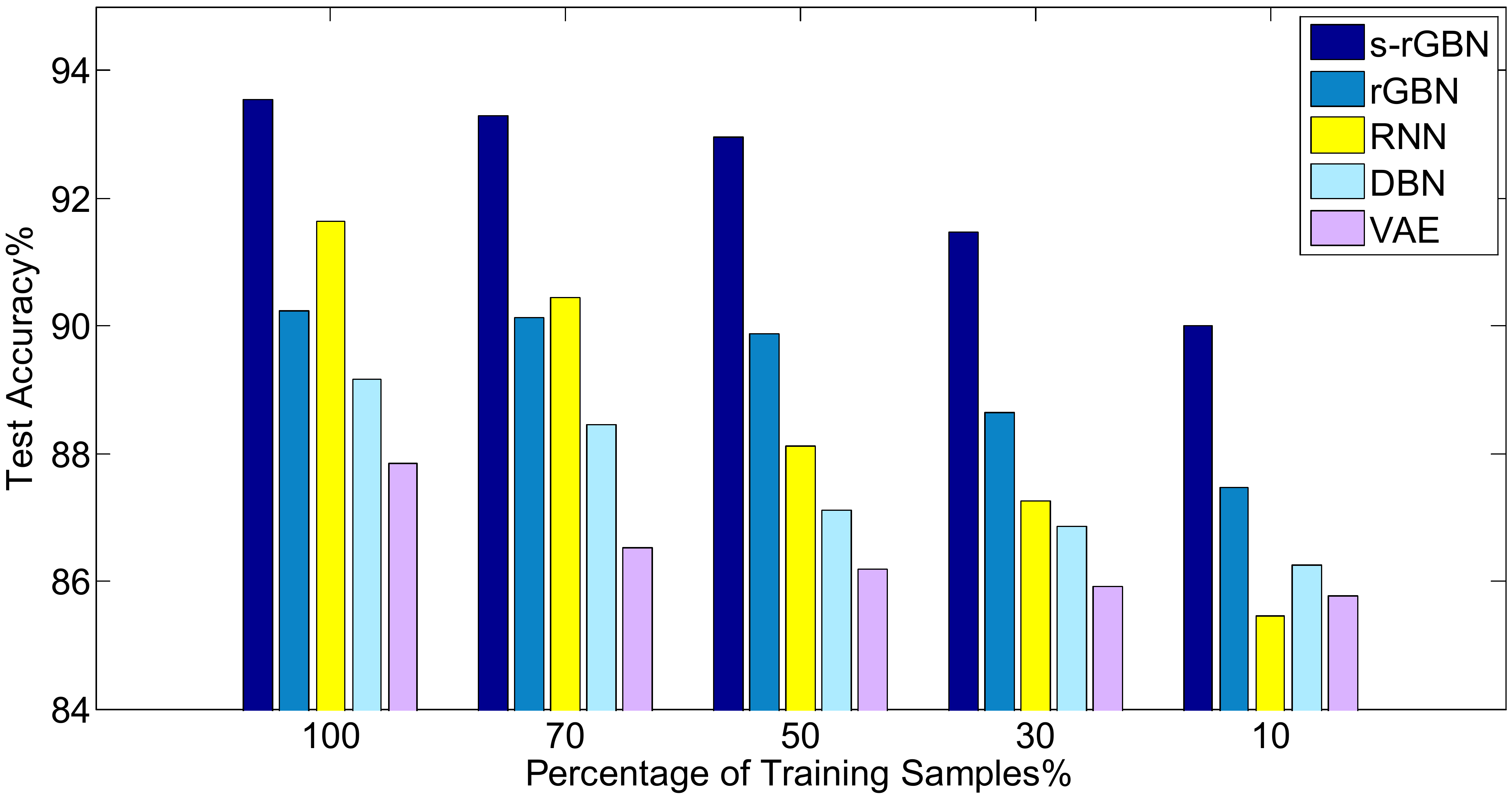}
\caption{Comparison of the recognition performance between various methods with different HPPR training set sizes.} 
\label{fig:trainingsize}
\end{center}
\end{figure}
\vspace{-0.1cm}
It is worth pointing that a practical RATR system should provide an acceptable recognition rate even with a few training samples~\cite{feng2017radar,WangLDS}, such as in the non-cooperative circumstance. In Fig. \ref{fig:trainingsize}, we depict how the recognition performance of each method varies with the size of the training dataset.
With a relatively small training dataset, the accuracy of deterministic RNN drops sharply; a possible explanation for this phenomenon is
that a point estimate by SGD ignores model uncertainty and the kind of variability observed in highly structured data is poorly modeled by the output probabilistic model alone. The parameters in the generative model of VAE (the weights) are updated by SGD towards a point estimate, leading to an obvious decrease in test accuracy.
The DBN is robust to the small training samples whose weights are updated with Gibbs sampling, but it does not efficiently incorporate both between-layer and temporal dependencies of the observed HRRP samples, leading to lower accuracy than that of the proposed methods.
By contrast, by exploiting temporal correlations, the proposed models achieve higher accuracy and maintain acceptable performance even when the training data size becomes much smaller.
We attribute that to the following three advantages: 1) building the novel dynamical probabilistic model to describe the complicated sequential HRRP samples;
2) developing the hybrid stochastic gradient MCMC and recurrent variational inference to update model parameters and provide model uncertainty;
3) fusing multi-stochastic-layer features to enhance robustness.

\subsection{Computational Complexity}
\begin{figure}[!htbt]
\vspace{-0.2cm}
\setlength{\abovecaptionskip}{-0.3cm}
\setlength{\belowcaptionskip}{-0.5cm}
\begin{center}
\centering
\includegraphics[height=3.5cm,width=4.2cm]{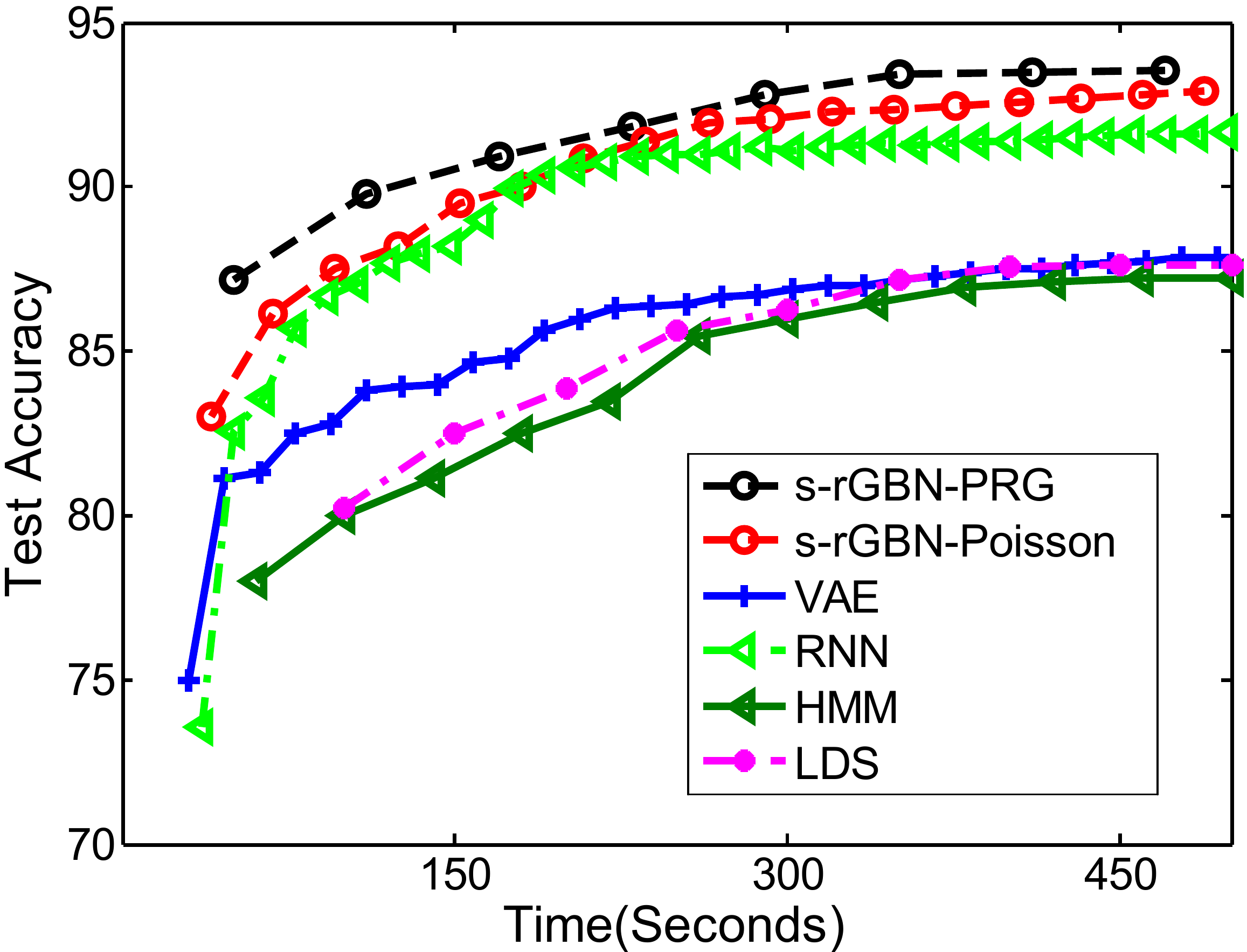}
\caption{Comparison of the test recognition performance as a function of training time between various methods.
}
\label{fig:training_time}
\end{center}
\end{figure}

\vspace{-0.3cm}
\begin{table}[!htbp]
\setlength{\abovecaptionskip}{-4pt}
\setlength{\belowcaptionskip}{-4pt}
 \centering
 \caption{Comparison of computational cost at the testing stage.}
 \scalebox{0.8}{\begin{tabular}{cc}
 \toprule
 \textbf{Models} &\textbf{Testing a sample(s)} \\
 \hline
 LDS & 0.6035 \\
 HMM & 0.4412\\
 VAE & 0.0006\\
 RNN layer-3 & 0.0008\\
 s-rGBN-Poisson layer-3 & 0.0010\\
 s-rGBN-PRG layer-3 &0.0010 \\
 \bottomrule
 \end{tabular}\label{Tab:Time}}
\end{table}

In this section, we evaluate the computational complexity of various methods.
In Fig. \ref{fig:training_time}, we compare various methods in terms of how their test recognition rates vary as the increase of training time.
At each training iteration, for both LDS and HMM, all training HRRP samples are processed, while for RNN, VAE, s-rGBN-PRG, and s-rGBN-Poisson, a mini-batch is randomly selected for training. As shown in Fig. \ref{fig:training_time}, both s-rGBN-PRG and s-rGBN-Poisson outperform RNN and VAE in providing higher performance as time progresses. Note that RNN is only a deterministic recurrent network that does not provide a probabilistic generative model in the latent space to model uncertainty and discover latent hierarchical structure, and VAE is restricted to model non-sequential data. Our mini-batch based inference algorithm converges faster than batch-based ones (including LDS and HMM), which further demonstrates the advantage of our proposed hybrid MCMC/VAE inference algorithm.

In Table \ref{Tab:Time}, we compare the computational complexity of various algorithms at the testing stage. Both HMM and LDS have high computational cost, due to the need to perform a number of iterations to infer the latent representation of each text sample. By contrast, by directly mapping a test data into its latent representation via non-linear transformation, RNN, VAE, and the proposed models equipped with the feed-forward inference network, are able to process the test data in real time.

\section{Conclusion}
For radar high-resolution range profile (HRRP) target recognition, we  introduce  recurrent gamma belief network (rGBN), a temporal deep generative model, to
efficiently capture the global structure of the targets and temporal dependence between the range cells in a single HRRP.
Scalable inference for rGBN is developed by integrating stochastic gradient MCMC
and recurrent variational inference into a hybrid inference scheme.
We further propose supervised rGBN to increase the discriminative power of the latent states by
jointly modeling the HRRP samples and their labels.
Experimental results on synthetic and measured HRRP data demonstrate that in comparison to existing models, the proposed ones not only exhibit
superior recognition performance and enhanced robustness to the variation of the training set size, but also provide highly interpretable latent structure.

%

%


\normalem
\bibliography{tsp_rGBN_RNN}
\bibliographystyle{IEEEtran}

\end{document}